\documentclass[10pt,twocolumn,letterpaper]{article}

\usepackage[pagenumbers]{cvpr} 

\usepackage{graphicx}
\usepackage{amsmath}
\usepackage{amssymb}
\usepackage{booktabs}

\usepackage{color}
\usepackage{algorithm,algorithmicx,algpseudocode}
\usepackage{bm,xspace}
\usepackage{comment}
\usepackage{bbold}
\usepackage{verbatim}
\usepackage{balance}
\usepackage{etoolbox,siunitx}
\usepackage{calc}
\usepackage{pifont,hologo}
\usepackage{adjustbox}
\usepackage{tabularx}
\usepackage{multirow}
\usepackage{caption} 
\usepackage{overpic}
\usepackage{array,makecell} 
\usepackage{esvect} 
\usepackage{textcomp, gensymb} 
\usepackage{tablefootnote}

\usepackage{xcolor,colortbl}
\definecolor{mygray}{gray}{0.9} 
\definecolor{myskyblue}{RGB}{230, 249,255}
\definecolor{myyellowgreen}{RGB}{240, 240, 245}

\newcolumntype{?}[1]{!{\vrule width #1}} 

\newcolumntype{Y}{>{\centering\arraybackslash}X}

\newcommand*{\myqcrfont}{\fontfamily{<qcr>}\selectfont}

\def\ourname{RIAV-MVS} 

\def\ie{\emph{i.e.}}

\usepackage[pagebackref,breaklinks,colorlinks]{hyperref}

\usepackage[capitalize]{cleveref}
\crefname{section}{Sec.}{Secs.}
\Crefname{section}{Section}{Sections}
\Crefname{table}{Table}{Tables}
\crefname{table}{Tab.}{Tabs.}

\def\cvprPaperID{4440} 
\def\confName{CVPR}
\def\confYear{2023}

\begin{document}

\title{RIAV-MVS: Recurrent-Indexing an Asymmetric Volume for Multi-View Stereo} 



\author{%
 Changjiang Cai, Pan Ji, Qingan Yan, Yi Xu \\
 OPPO US Research Center, InnoPeak Technology, Inc. \\
}


\maketitle

\begin{abstract}
This paper presents a learning-based method for multi-view depth estimation from posed images. Our core idea is a ``learning-to-optimize'' paradigm that iteratively indexes a plane-sweeping cost volume and regresses the depth map via a convolutional Gated Recurrent Unit (GRU). Since the cost volume plays a paramount role in encoding the multi-view geometry, we aim to improve its construction both at pixel- and frame- levels. At the pixel level, we propose to break the symmetry of the Siamese network (which is typically used in MVS to extract image features) by introducing a transformer block to the reference image (but not to the source images). Such an asymmetric volume allows the network to extract global features from the reference image to predict its depth map. Given potential inaccuracies in the poses between reference and source images, we propose to incorporate a residual pose network to correct the relative poses. This essentially rectifies the cost volume at the frame level. We conduct extensive experiments on real-world MVS datasets and show that our method achieves state-of-the-art performance in terms of both within-dataset evaluation and cross-dataset generalization. Code available: \url{https://github.com/oppo-us-research/riav-mvs}.
\end{abstract}

\section{Introduction}
\label{sec:intro}
\begin{figure*}[!htb]
    \centering
    \renewcommand{\tabcolsep}{1pt}
    \scriptsize
    \includegraphics[width=0.9\textwidth]{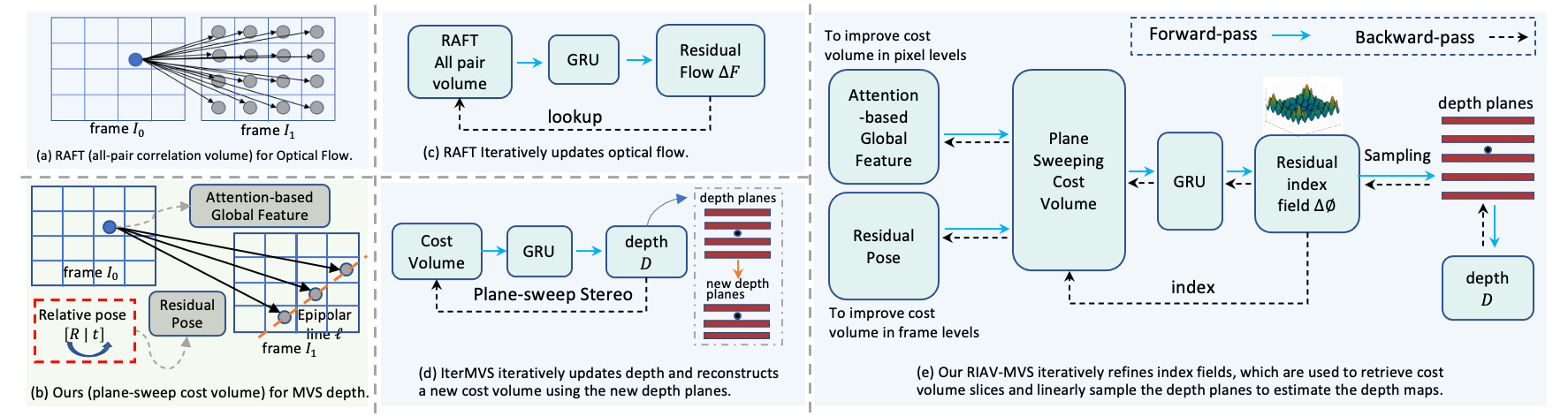}
    \caption{Our pipeline versus RAFT~\cite{raft-teed2020} and IterMVS~\cite{itermvs_wang2021}. Our recurrent processing of a plane-sweep cost volume by the iteratively refined index field serves as a new design for multi-view depth estimation.}
    \label{fig:our-vs-raft-comparison}
    \vspace{-8pt}
\end{figure*}

Multi-view stereo (MVS) aims to recover dense 3D geometry from multiple images captured from different viewpoints with calibrated cameras \cite{surfacenet_ji2017}. It is a fundamental problem in computer vision and has wide applications ranging from autonomous driving \cite{Chen_2017_CVPR_mvs_autonomous_driving,Wang_2019_CVPR_Pseudo_LiDAR}, remote sensing \cite{agarwal2011building_rome}, augmented reality \cite{google-depth-from-smartphone-AR}, to robotics \cite{hane2011stereo}.  
Following the seminal MVSNet \cite{yao2018mvsnet}, many learning-based methods~\cite{deepvideomvs2021,Long_2021_CVPR_EstDepth,luo2019pmvsnet,itermvs_wang2021,patchmatchnet_wang2020,Yang_2020_CVPR_pyramid_cost,yao2019recurrent} have been proposed, achieving great improvements against their traditional counterparts~\cite{bleyer2011patchmatch,space-sweep-collins,gallup2007real,schoenberger2016sfm}, in terms of accuracy or efficiency. 

Most of the learning-based MVS methods \cite{deepvideomvs2021,Long_2021_CVPR_EstDepth,luo2019pmvsnet,itermvs_wang2021,Yang_2020_CVPR_pyramid_cost,yao2019recurrent} rely on traditional plane-sweeping \cite{space-sweep-collins,gallup2007real} approach to generate a cost volume by comparing the CNN features of reference image and source images at several depth hypotheses, and then apply 2D or 3D convolutional encoder-decoders to aggregate and regularize the cost volume. The 2D CNN methods \cite{deepvideomvs2021} use multi-level features as skip connections to help decode the cost volume for depth regression. Even though the skip connections improve the depth maps, they weaken the role of cost volume and the geometry knowledge embedded therein to some extent.  Hence, 2D CNN methods suffer from degraded generalization when testing on unseen domains. The 3D CNN methods \cite{kendall2017-gcnet} use \textit{soft-argmin} to regress the depth map as the expectation from the cost volume distribution, and hence cannot predict the best candidate but instead an averaged one when dealing with a flat or multi-modal distribution caused by textureless, repeated, or occluded regions, etc. To mitigate these problems, we propose \ourname, a new paradigm to predict the depth via learning to recurrently index an asymmetric cost volume, obtaining improved accuracy and generalization. As depicted in Fig. \ref{fig:our-vs-raft-comparison}, our \ourname{} features several nontrivial novel designs. 

First, we learn to index the cost volume by approaching the correct depth planes per pixel via an \textit{index field} (a grid of indices to identify the depth hypotheses), as shown in Fig.~\ref{fig:our-vs-raft-comparison}-(e). The proposed recurrent estimate of the index field  enables the learning to be anchored at the cost volume domain. Specifically, it recurrently predicts the residual index field in a descent direction of matching cost 
to retrieve cost values for the next iteration. The newly updated index field is used to directly index (\ie, sampling via linear interpolation) depth hypotheses to render a depth map, which is iteratively optimized to approach the ground truth depth, making the system end-to-end trainable.

Second, to facilitate the optimization, we propose to improve the cost volume at pixel- and frame- levels, respectively. At the pixel level, a transformer block is asymmetrically applied to the reference view (but not to the source views). By capturing long-range global context via a transformer and pixel-wise local features via CNNs, we build an asymmetric cost volume to store more accurate matching similarity cues. At the frame level, we propose a residual pose net to rectify the camera poses that are usually obtained via Visual SLAM~\cite{campos2021orb,dai2017bundlefusion,ji2022georefine} and inevitably contain noise. The rectified poses are used to more accurately backward warp the reference features to match the counterparts in source views. 

Our \ourname{} is depicted versus two related works RAFT~\cite{raft-teed2020} and IterMVS~\cite{itermvs_wang2021} as in Fig.~\ref{fig:our-vs-raft-comparison}. First, our method is developed using RAFT's GRU-based iterative optimization. However, RAFT operates an all-pair correlation volume (no multi-view geometry constraints) for optical flow (Fig.~\ref{fig:our-vs-raft-comparison}-(a) and (c)), our method is proposed for multi-view depth estimation by constructing a plane-sweep cost volume (Fig.~\ref{fig:our-vs-raft-comparison}-(b)). Second, IterMVS~\cite{itermvs_wang2021} iteratively predicts the depth and reconstructs a new plane-sweep cost volume using updated depth planes centered at the predicted depth (Fig.~\ref{fig:our-vs-raft-comparison}-(d)). Instead, as shown in Fig.~\ref{fig:our-vs-raft-comparison}-(e), our proposed \textit{index field} serves as a new design that bridges the cost volume optimization (\ie, by learning better image features via back-propagation) and the depth map estimation (\ie, by sampling sweeping planes). It makes forward and backward learning differentiable. We conduct extensive experiments on indoor-scene datasets, including ScanNet~\cite{dai2017scannet}, DTU~\cite{dtu_jensen2014large}, 7-Scenes~\cite{7scenes-glocker2013real-time}, and RGB-D Scenes V2~\cite{lai2014unsupervised}. We also performed well-designed ablation studies to verify the effectiveness and the generalization of our approach.

\section{Related Work}\label{sec:related}
Depth can be accurately predicted from stereo matching, which can be broadly divided into binocular stereo and multi-view stereo (MVS). The former requires calibrated setups of rectified stereo pairs, and many traditional \cite{batsos2018cbmv,hirschmuller08,hu2012quantitative,scharstein2002taxonomy} and deep learning-based methods \cite{cai2020deep_adaptive_stereo,cai2020matchingspace,chang2018psmnet,kendall2017-gcnet,Li2022PracticalSM,liang2018learning_1st_Rob,mayer2016large,zbontar2016stereo,zhang2019ga} have been proposed. Compared with binocular stereo, MVS methods estimate depth from a set of images or a video, where the camera moves and the scene is assumed static. 
In this section, we briefly review deep learning-based MVS methods.

\vspace{1pt}
\noindent \textbf{3D-CNN MVS Depth Estimation:} \,\, Learning-based MVS methods \cite{deepvideomvs2021,Long_2021_CVPR_EstDepth,luo2019pmvsnet,itermvs_wang2021,patchmatchnet_wang2020,Yang_2020_CVPR_pyramid_cost,yao2018mvsnet,yao2019recurrent} rely on traditional plane-sweeping \cite{space-sweep-collins,gallup2007real} to generate a cost volume by associating reference frame and source frames for similarity matching, followed by encoder-decoder architectures for cost volume aggregation and depth map prediction. Among them, MVSNet \cite{yao2018mvsnet}, R-MVSNet \cite{yao2019recurrent}, and DPSNet \cite{im2019dpsnet} leverage 3D convolutions to regularize 4D cost volumes and regress depth maps via \textit{soft-argmin} \cite{kendall2017-gcnet}. Different strategies have been introduced for cost volume construction. MVSNet \cite{yao2018mvsnet} proposes a variance-based cost volume for multi-view similarity measurement. Cas-MVSNet \cite{Gu_2020_CVPR_Cascade_Cost_Volume} builds cascade cost volumes based on multi-scale feature pyramid and regresses the depth map in a multi-stage coarse-to-fine manner. Similarly, Cheng \etal propose UCS-Net \cite{Cheng_2020_CVPR_Adaptive_Thin} to build cascade adaptive thin volumes by leveraging variance-based uncertainty. CVP-MVSNet \cite{Yang_2020_CVPR_pyramid_cost} builds a cost volume pyramid via multi-scale images to reduce to memory footprint. This cost volume mechanism is also adopted to other related tasks, \eg, 3D plane reconstruction in PlaneMVS \cite{liu2022planemvs}, which leverages slanted plane sweeping to help plane reconstruction and accurate depth predictions.

\vspace{1pt}
\noindent \textbf{2D-CNN MVS Depth Estimation:} \,\, Even though 3D convolutional methods usually deliver high accuracy, they demand a large memory footprint and computational cost. Instead, some methods, \eg, MVDepthNet \cite{MVDepthNet2018} and DeepVideoMVS \cite{deepvideomvs2021}, generate 3D volumes by computing correlation or dot production between the extracted features of multi-view input images. The 3D cost volumes are further regularized by 2D convolutions. DeepVideoMVS \cite{deepvideomvs2021} extracts multi-scale features. It uses the feature at half scale to construct the cost volume, and other features as the skip connections to a series of decoders for depth regression. PatchmatchNet \cite{patchmatchnet_wang2020} proposes an adaptive procedure mimicking PatchMatch \cite{bleyer2011patchmatch} to achieve superior efficiency. 2D convolutions are much faster and more memory efficient than the 3D counterparts, making them better suitable for lightweight networks in real-time applications. 

\vspace{1pt}
\noindent \textbf{Iterative Depth Estimation:} \,\,  Several methods adopte an iterative depth estimation paradigm. R-MVSNet \cite{yao2019recurrent} iteratively regularizes each slice of the cost volume with GRU along the depth dimension. Unlike the above mentioned cost volume-based approaches, Point-MVSNet \cite{chen2019point_mvs} directly processes the target scene as point clouds. It first generates a coarse depth map, converts it into a point cloud and refines the point cloud iteratively to reduce the residual between the estimated depth and the ground truth depth. IterMVS \cite{itermvs_wang2021} encodes a pixel-wise probability distribution of depth in the hidden state of a GRU-based estimator. During each iteration, the multi-scale matching information is injected into the GRU to predict the depth and confidence maps to facilitate the following 3D reconstruction. The depth maps are predicted via a combined classification and regression through the probability distribution. It uses \textit{arg-min} to finish ``classification'', which is not differentiable and must be detached first before the following regression operation. Unlike IterMVS, our method learns to recurrently index the cost volume to directly find the best depth candidates in an end-to-end differentiable fashion. 

\section{Method}\label{sec:method}
Our learning-based end-to-end multi-view stereo system, \ourname, aims to predict depth maps from a set of images, which are different views of the same scene with known camera poses, denoted by $\mathcal{I} =\{I_i\}_{i=0}^{N-1}$. More specifically, \ourname~ uses one as the reference image and others as source images to infer the depth map of the reference image. Without loss of generality, we refer to the first image $I_0$ as the reference image and others as source views $\mathcal{I}^\mathcal{S}$, where $S={1,2,...,N-1}$. An overview of our approach is illustrated in Fig. \ref{fig:overview-archi}. It consists of feature extraction (Sec. \ref{method:feature}), cost volume construction via plane-sweeping stereo \cite{space-sweep-collins,gallup2007real} (Sec. \ref{method:cost}), and cost volume optimization and depth prediction (Sec. \ref{method:iterative}). 
Details will be discussed below.


\subsection{Feature Extraction} \label{method:feature}

\begin{figure*}[!t]
    \centering
    \renewcommand{\tabcolsep}{1pt}
    \scriptsize
    \includegraphics[width=0.9\textwidth]{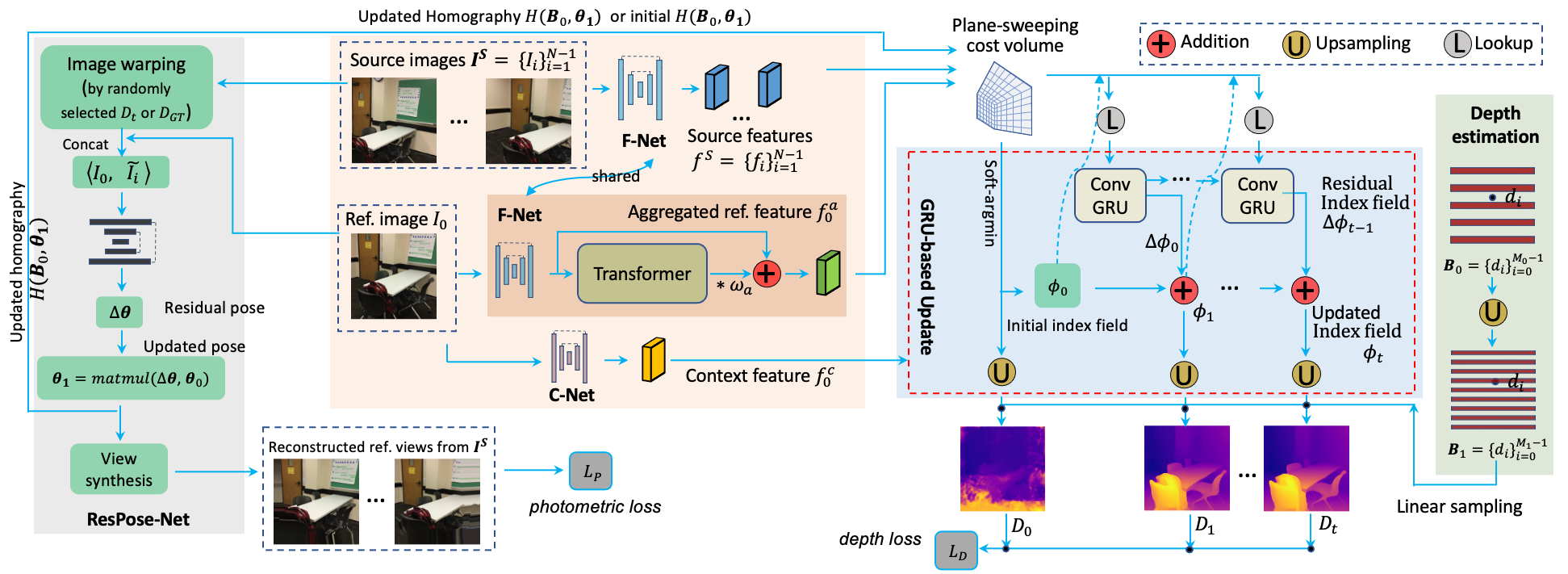}
    \caption{Architecture of our proposed network. It consists of a feature extraction (\ie, F-Net, a Transformer, and C-Net) block, a cost volume construction and index field GRU-based optimization block, and a residual pose block.}
    \label{fig:overview-archi}
\end{figure*}

Given a reference image $I_0$ and source images $\mathcal{I}^\mathcal{S}$, we extract matching features of $I_0$ and each of the $\mathcal{I}^\mathcal{S}$ by \textit{F-Net} (see below), and a context feature for $I_0$ by \textit{C-Net} (see the supplementary material for \textit{C-Net}), as shown in Fig. \ref{fig:overview-archi}.

\noindent \textbf{Local Matching Feature Extraction:} \,\, Our feature extractor F-Net is based on PairNet \cite{deepvideomvs2021}. It is a lightweight Feature Pyramid Network (FPN) \cite{fpn_Lin_2017} on top of the first fourteen layers of MnasNet \cite{mnasnet_tan2019}. Specifically, the reference input image $I_0 \in \mathbb{R} ^{H \times W}$ is spatially scaled down until 1/32 scale, and recovered up to 1/2 scale, resulting in multi-scale features $\{f_{0,s} \in \mathbb{R} ^{\frac{H}{s} \times \frac{W}{s} \times F_0} \}$ ($s$=2,4,8,16 and $F_0$=32 for feature channels). In PairNet, $f_{0,2}$ is used to construct the cost volume, and other features are used in the skip connections to a series of decoders for depth regression. Unlike this, we add an extra fusion layer $\mathcal{G}$ to aggregate them into a matching feature $f_0$ at 1/4 scale, as
\vspace{-5pt}
\begin{equation}\label{eq:feature_fusion}
f_0 = \mathcal{G}(\langle  f_{0,2}\downarrow _2, f_{0,4}, f_{0,8}\uparrow _2,  f_{0,16}\uparrow _4 \rangle) 
\end{equation}

\noindent where the fusion layer $\mathcal{G}$ is a sequence of operations of $\text{Conv}_{3 \times 3}$, batch normalization, ReLU, and $\text{Conv}_{1 \times 1}$, $\downarrow _x $ and $\uparrow _x $ are downsampling and upsampling by scale $x$, $\langle \cdot \rangle$ is concatenation along channel dimension, and $f_0 \in  \mathbb{R} ^{H/4 \times W/4 \times F_1}$ with $F_1$=128. Similarly, F-Net (with shared weights as that for $I_0$) is also applied to source images $\mathcal{I}^\mathcal{S}$ to extract a set of matching features ${f}^{\mathcal{S}} = \{f_{i} \mid i \in \mathcal{S} \}$. 

\noindent \textbf{Global Matching Feature of Reference View:} \,\, Besides the local pixel-wise features extracted from CNNs, we also leverage global long-range information to better guide the feature matching. Towards that, a transformer layer (four-head self-attention with positional encoding) \cite{attention_Vaswani2017} is applied to the local feature $f_0$ of the reference image, to construct an aggregated feature $f_0^a \in  \mathbb{R}^ {H/4 \times W/4 \times F_1}$ as 


\vspace{-8pt}
\begin{equation}\label{eq:transformer_globalfeat}
f_0^a = f_0 +  \omega_\alpha \sigma \left( 
\frac{(f_0 W^Q) (f_0 W^K)^T}{\sqrt{F_1}} (f_0 W^V )
\right)
\end{equation}
\vspace{-8pt}

\noindent where $\sigma(\cdot)$ is the softmax operator, $\omega_\alpha$ is a learned scalar weight that is initialized to zero, and $W^Q$, $W^K \in \mathbb{R}^ {F_1 \times hF_1}$, and $W^V \in  \mathbb{R}^ {F_1 \times F_1}$are the projections matrices for query, key and value features, with $h$=4 for multi-head attention. 
The final output $f_0^a$ contains both local and global information, which are balanced by the parameter $\omega_\alpha$, to enhance the following cost volume construction. 

It is worth noting that this transformer self-attention is only applied to the \textit{reference image}, while the source features still possess the local representations from CNNs. Our \textit{asymmetric} employment of this transformer layer provides the capability to better balance the high-frequency features (by high-pass CNNs) and the low-frequency features by self-attention \cite{park2022how, craft_Sui_2022}. The high-frequency features are beneficial to image matching at local and structural regions, while the low-frequency ones, with noisy information suppressed by the transformer's spatial smoothing (serving as a low-pass filter), provide more global context cues for robust matching, especially for the areas full of low-texture, repeated patterns, and occlusion, etc. This way, our network can learn where to rely on global features over local features, and vice versa.

\subsection{Cost Volume Construction} \label{method:cost}

We use the global matching feature map $f_0^a$ of $I_0$ and local matching features ${f}^{\mathcal{S}}$ of $\mathcal{I}^\mathcal{S}$ to build a cost volume. 
The cost (or matching) volume is defined on a 3D view frustum attached to the camera in perspective projection. It is generated by running the traditional plane-sweep stereo \cite{space-sweep-collins, gallup2007real} which uniformly samples $M_0$=64 plane hypotheses in the inverse depth space, s.t. $ 1/d \sim U(d_{\text{min}}, d_{\text{max}})$. Here $d_{\text{min}}$ and $d_{\text{max}}$ are the near and far planes of the 3D frustum, respectively. Following \cite{deepvideomvs2021}, we set $d_{min}$=0.25 and $d_{max}$=20 meters for indoor scenes (e.g., ScanNet \cite{dai2017scannet}). 

For a given depth hypothesis $d$ and known camera intrinsic matrices $\mathcal{K} = \{K_i\}_{i=0}^N$ and relative transformations $\Theta = \{R_{0,i} \mid  t_{0,i} \}_{i=1}^N$ from reference image $I_0$, to source image $I_i$, a cost map is computed by i) warping source image feature $f_i$ into the reference image and ii) calculating the similarity between the reference global feature $f_0^a$ and the warped feature $\tilde{f}_i$. To generate $\tilde{f}_i$, we implement the homography ${H}$ as a backward 2D grid sampling. Specifically, a pixel $\mathbf{p}=(u,v, 1)^T$ in the reference image will be warped to its counterpart $\tilde{\mathbf{p}}_i$ in source image $i$ as follows:

\vspace{-5pt}
\begin{equation}\label{eq:Homography}
\tilde{\mathbf{p}}_i = H(\mathbf{p} \mid d, \mathcal{K}, \Theta) = {K}_i \left( {R}_{0,i} \left({K}_{0}^{-1} \mathbf{p}  d \right)  + t_{0,i} \right )
\end{equation}
\vspace{-4pt}

Then $\tilde{f}_i$ is bilinearly sampled from $f_i$ as $\tilde{f}_i (\mathbf{p}) = {f}_i (\tilde{\mathbf{p}}_i)$. Given the warped source feature $\tilde{f}_i$ and the reference feature $f_0^a$, the cost volume is formulated as $C_0 (d) = \frac{1}{N-1} \sum_{i \in \mathcal{S}} \frac{f_0^a \cdot  \tilde{f}_i^T}{\sqrt{F_1}}$. This way, we can construct a cost volume for all depth candidates $\mathcal{B}_0$=$\{ d_i\}_{i=0}^{M_0-1}$, resulting in a 3D tensor, denoted as $C_0 \in  \mathbb{R}^ {H/4 \times W/4 \times M_0}$.

\subsection{GRU-based Iterative Optimization} \label{method:iterative}


We solve the depth prediction as learning to optimize the dense stereo matching problem \cite{bleyer2011patchmatch,hirschmuller08,taniai2018local-expansion,xu2020marmvs}. Given the generated cost volume $C_0$ as in Sec.~\ref{method:cost}, the depth estimation of the reference image is formulated as finding the best solution $D^* = \operatorname {argmin} _{D} \,E(D, C_0)$, which minimizes an energy function $E(D, C_0)$ (including a data term and a smoothness term). Unfortunately, such a \textit{global} minimization is NP-complete due to many discontinuity preserving energies \cite{boykov2001fast}.  Approximate solutions are proposed by loosening 
the energy function, \eg, the binocular stereo matching solved by Semi-Global Matching (SGM)\cite{hirschmuller08}. In SGM, the matching cost $C_0$ is iteratively aggregated by summing the costs (of all 1D minimum cost paths that end in
pixel $\mathbf{p}$ at disparity $d$\footnote{Abuse of notation $d$ for depth in MVS or disparity in binocular stereo}) when traversing from pixel $\mathbf{p} - \mathbf{r}$ to pixel $\mathbf{p}$ in a direction $\mathbf{r}$ (out of sixteen directions) and the best disparity at each pixel $\mathbf{p}$ is given by ${ d^{*}(\mathbf{p})=\operatorname {argmin} _{d} (C^\prime(\mathbf{p},d))}$, with $C^\prime$ being the aggregated cost volume. Similar to SGM, we do not directly optimize the energy function $E$, but learn to process the cost volume $C_0$.

However, several major problems still need to be solved. SGM is not differentiable due to its winner-take-all (WTA) by $\operatorname {argmin} $, making it unable to train the system in an end-to-end manner. Its differentiable counterpart, SGA \cite{zhang2019ga}, was proposed by changing the $\operatorname {min} $ to $\operatorname {sum} $ when aggregating the cost volume, and replacing the $\operatorname {argmin} $ with $\operatorname {softargmin} $ when predicting the disparity from the optimized cost volume, but still i) the update direction $\mathbf{r}$ when traversing from pixel $\mathbf{p}$-$\mathbf{r}$ to $\mathbf{p}$ needs to be predefined, and ii) the $\operatorname {softargmin} $ focuses on measuring the distance of the \textit{expectation} of disparity map to the ground truth disparity, and hence cannot handle multi-modal distributions in $C_0$ well 
\cite{itermvs_wang2021}. 

Therefore, towards an end-to-end, differentiable solution, we propose to use a GRU-based module to implicitly optimize the matching volume. It estimates a sequence of \textit{index fields} $\{ \phi _t \}_{t=1}^{T}$ by unrolling the optimization problem to $T$ iterative updates (in a descent direction), mimicking the updates of a first-order optimizer according to \cite{adler2017solving,adler2018learned,raft-stereo-lipson2021,raft-teed2020}. At each iteration $t$, the index field $\phi_t \in \mathbb{R} ^{H \times W} $ is estimated as a grid of indices to iteratively better approach (\ie, closer to the ground truth) depth hypotheses having a lower matching cost. Specifically, a \textit{residual} index field $\delta \phi_t$ is predicted as an update direction for next iteration, \ie, $\phi_{t+1} = \phi_t  + \delta \phi _t$, (analogous to the direction $\mathbf {r}$ in SGM), which is explicitly driven by training the system (\eg, feature encoders, the transformer layer, and the residual pose net, etc.) to minimize the loss between the predicted depth maps and the ground truth. The recurrent estimate of the index field enables the learning to be directly anchored at the cost volume domain.
This indexing paradigm differentiates our approach from other depth estimation methods, such as the depth regression which fuses cost volume and the skipped multi-level features by 2D CNNs \cite{deepvideomvs2021,patchmatchnet_wang2020,itermvs_wang2021}, and soft-argmin \cite{kendall2017-gcnet} after cost volume aggregation and regularization by 3D CNNs \cite{Gu_2020_CVPR_Cascade_Cost_Volume, Long_2021_CVPR_EstDepth, yao2018mvsnet,yao2019recurrent}. 




\noindent \textbf{Index Field Iterative Updates:} \,\, We use a 3 chained GRUs \cite{raft-stereo-lipson2021} to estimate a sequence of index fields, $\{ \phi _t \in \mathbb{R} ^{H/4 \times W/4} \}_{t=1}^{T}$ from an initial starting point $\phi _0$. 
We use a \textit{softargmin-start} from the cost volume $C_0$, \ie, $ \phi_0 = \sum_{i=0}^{M_1-1} i \sigma(C_0)$, where $\sigma(\cdot)$ is the $\operatorname {softmax}$ operator along the last dimension of cost volume $C_0$, to convert it to a probability of each index $i$. This setup facilitates the convergence of our predictions. 
A four-layer matching pyramid $\{ C_0^i \in \mathbb{R} ^{H/4 \times W/4 \times M_0/{2^i}} \}_{i=1}^4$ is built by repeated pooling the cost volume $C_0$ along the depth dimension with kernel size $2$ as in \cite{raft-teed2020}. To index the matching pyramid, we define a lookup operator 
analogous to the one in \cite{raft-stereo-lipson2021}. Given a current estimate of index field $\phi_t$, a 1D grid is constructed with integer offsets up to $r=\pm4$ around the $\phi_t$. The grid is used to index from each level of the matching pyramid via linear interpolation due to $\phi_t$ being real numbers. The retrieved cost values are then concatenated into a single feature map $C_0^{\phi_t} \in \mathbb{R} ^{H/4 \times W/4 \times F_2}$. 
Then the index field $\phi_t$, the retrieved cost features $C_0^{\phi_t}$, and context features $f_0^c$ are concatenated, and fed into the GRU layer, together with a latent hidden state $h_{t}$. The GRU outputs a residual index field $\delta \phi_{t}$, and a new hidden state $h_{t+1}$: 

\vspace{-8pt}
\begin{equation*}\label{eq:gru_update}
\delta \phi_t, h_{t+1} \Leftarrow  \operatorname{GRU}(\langle \phi_t, C_0^{\phi_t}, f_0^c \rangle, h_{t}); \phi_{t+1} \Leftarrow \phi_{t} + \delta \phi_t 
\end{equation*}


\noindent \textbf{Upsampling and Depth Estimation:} \,\, The depth map at iteration $t$ is estimated by sampling the depth hypotheses via linear interpolation given the index field $\phi_t$. Since $\phi_t$ is at 1/4 resolution, we upsample it to full resolution using a convex combination of a 3$\times$3 neighbors as in \cite{raft-teed2020}. Specifically, a weight mask $W_0 \in \mathbb{R} ^ {H/4 \times W/4 \times (4 \times 4 \times 9) }$ is predicted from the hidden state $h_t$ using two convolutional layers and $\operatorname {softmax}$ is performed over the weights of those 9 neighbors. The final high resolution index field $\phi_t^{u}$ is obtained by taking a weighted combination over the 9 neighbors, and reshaping to the resolution H$\times $W. Convex combination can be implemented using the \textit{einsum} function 
in PyTorch. 

When constructing the cost volume, we use $M_0=64$ depth hypotheses, $\mathcal{B}_0=\{d_i\}_{i=0}^{M_0 -1 }$. A small $M_0$ helps reduce the computation and space. If we use the upsampled index field $\phi_t^{u}$ to directly sample the planes $\mathcal{B}_0$, we see \textit{discontinuities} in the inferred depth map, even though the quantitative evaluation is not hindered. To mitigate this, we propose a coarse-to-fine pattern, and to use $M_1$=256 depth hypotheses $\mathcal{B}_1=\{d_i\}_{i=0}^{M_1 -1 }$. Analogous to upsampling in optical flow or disparity in binocular stereo, the flow or disparity values themselves have to be scaled when implementing the spatial upsampling. Our depth index fields are adjusted by a scale $s_D$=$\frac{M_1}{M_0}$=4. To mimic the convex combination before mentioned, we apply a similar weighted summation along the depth dimension when sampling depth from $\mathcal{B}_1$. Specifically, another mask $W_1 \in \mathbb{R} ^ {H \times W \times s_D \times M_0 }$ is predicted from the hidden state using three convolutional layers, and further reshaped to $H \times W \times M_1$. Given a pixel $\mathbf{p}$, and the upsampled index field $\phi_t^{u}$, the final depth $D_t$ is estimated as


\vspace{-8pt}
\begin{equation}\label{eq:depth_upsample}
D_{t}(p) = \frac{\sum_{i \in \Omega(\mathbf{p})} \mathcal{B}_1 \left[ i \right]W_1(\mathbf{p}, \lfloor i \rfloor)}{\sum_{i \in \Omega(\mathbf{p})} W_1(\mathbf{p}, \lfloor i \rfloor)}
\end{equation}
\vspace{-8pt}
 

\noindent where, we aggregate the neighbors within a radius $r=4$ centered at the index $\phi_t^{u}(\mathbf{p})$ for a given pixel $\mathbf{p}$, and $\lfloor i \rfloor$ gives a greatest integer less than or equal to $i$, and $\left[ i \right]$ means to index the depth planes $\mathcal{B}_1$ via linear interpolation, due to index $i$ being a real number.

It is worth mentioning that our method embeds both regression (similar to \textit{softargmin} in existing methods \cite{Long_2021_CVPR_EstDepth,yao2018mvsnet, yao2019recurrent}) and classification (similar to \textit{argmin}), which make it robust to multi-modal distributions, and achieving \textit{sub-pixel} precision thanks to linear interpolation. Combining both classification and regression has been seen in \cite{itermvs_wang2021}, but ours does not use the \textit{argmax} operator when achieving the ``classification'' purpose, thanks to our proposal of index fields estimation, which differentiably bridges the cost volume indexing and depth hypotheses sampling directly in a sub-pixel precision.

\vspace{1pt}
\noindent \textbf{Residual Pose Net:} \,\, An accurate cost volume benefits the GRU-based iterative optimization. The quality of the generated cost volume $C_0$ is not only determined by the matching features ($f_0^a$ and $f^{ \mathcal{S}}$) (for which we have proposed asymmetric employment of the transformer layer), but also by the homography warping as in Eq. \ref{eq:Homography}. However, the camera poses are, in practice, usually obtained by Visual SLAM algorithms \cite{campos2021orb,dai2017bundlefusion,ji2022georefine}, and inevitably contain noise. Therefore, we propose a residual pose net to rectify the camera poses for accurately backward warping the reference features to match the corresponding features in source images. We use an image-net pretrained \textit{ResNet18} \cite{he2016deep} backbone as in \cite{manydepth_watson2021temporal,ji2021monoindoor} to encode the reference image and the warped source images. Specifically, given the current estimated depth map $D_t$ at iteration $t$, and the ground truth depth $D_{gt}$, we warp a source image $I_i$ into the reference image through the homography defined in Eq. \ref{eq:Homography} with (noisy) ground truth camera poses $\Theta$ and the $D_t$ or $D_{gt}$. We randomly select $D_t$ or $D_{gt}$ with a probability prob($D_t$)=0.6 during network training, but always use the predicted depth $D_t$ during network inference. The input to the pose net is the concatenated $I_0$ and the warped $\tilde{I}_i$, and the output is an axis-angle representation, which is further converted to a residual rotation matrix $\Delta \theta_i$, for an updated one $\theta^\prime_i$= $\Delta\theta_i \cdot \theta_i$. This way, we predict the residual poses $\Delta \Theta = \{ \Delta \theta_i \}_{i=1}^{N-1}$ for each of the source and reference pair, and perform the rectification as $\Theta^\prime = \Delta \Theta \cdot \Theta$. We leverage the updated poses $\Theta^\prime$ to calculate a more accurate cost volume $C_1$ using Eq. \ref{eq:Homography}, followed by the remaining iterations of GRU.


\subsection{Loss Function} \label{method:loss}
Our network is supervised on the inverse $L_1$ loss between the predicted depths $\{D_{t}\}_{i=1}^{T}$ and the ground truth $D_{gt}$. It is evaluated over valid pixels (\ie,
with non-zero ground truth depths). Following the exponentially increasing weights as in \cite{raft-stereo-lipson2021,raft-teed2020}, this depth loss is defined as 
\begin{equation}\label{eq:loss}
\mathcal{L}_{D} = \sum_{t=1}^{T} \gamma ^{T-t}  \frac{1}{N_v}  \sum_{i=0}^{N_v-1} \| \frac{1}{ D_t(i)} - \frac{1}{D_{gt}(i)} \|_1   
\end{equation}
\vspace{-2pt}

\noindent where, $\| \cdot \|_1$ measures the $l_1$ distance, $N_v$ is the number of valid pixels, and $\gamma=0.9$. We also apply the photometric loss $\mathcal{L}_{P}$(as defined in~\cite{manydepth_watson2021temporal}) to supervise the residual pose network. The total loss is then defined as $\mathcal{L} = \mathcal{L}_{D} + \mathcal{L}_{P}$.

\section{Experiments}\label{sec:experiments}
\subsection{Datasets}
Our experiments use four indoor-scene datasets, which have RGB-D video frames with ground truth depths and known camera poses. ScanNet \cite{dai2017scannet} and DTU \cite{dtu_jensen2014large} are used in training and testing, and 7scenes \cite{7scenes-glocker2013real-time} and RGB-D Scenes V2 \cite{lai2014unsupervised} are evaluated for zero-shot generalization. (1) \textbf{ScanNet}. Our network is trained from scratch on ScanNet \cite{dai2017scannet} using the official training split. We use 279k training samples and 20k validation ones. (2) \textbf{DTU.} 
Following \cite{patchmatchnet_wang2020,yao2018mvsnet,yao2019recurrent}, the depth range for sampling depth hypotheses is set to $d_{min} = 0.425$ and $d_{max} = 0.935$ meters. We use 27k training samples, 6k validation ones, and 1k ones for evaluation. Each sample has 5 frames. (3) \textbf{7-Scenes}. We select 13 sequences from 7-Scenes for zero-shot generalization. (4) \textbf{RGB-D Scenes V2}. We select 8 sequences for testing. Details about train/val/test splits, training, and implementation are shown in the \textit{supplementary material}.
 
 \subsection{Comparison with Existing Methods}
 
 In this section, our method is evaluated and compared with several state-of-the-art MVS methods. Our network is strictly compared with two baselines PairNet \cite{deepvideomvs2021} and IterMVS \cite{itermvs_wang2021}, following the same training schedule and training set. We also compare ours with other MVS methods either by running the provided models or referring to the available evaluation metrics when testing on the same datasets, including ESTDepth \cite{Long_2021_CVPR_EstDepth}, Neural RGBD \cite{liu2019neural}, MVDepthNet \cite{MVDepthNet2018}, DPSNet\cite{im2019dpsnet}, and DELTAS \cite{deltas-sinha2020}. 
 
 \begin{table*}[!tb]
\small
\vspace{1pt}
\centering
\scalebox{0.92}{
    \begin{tabularx}{0.98\linewidth}{ l | c c  c c  c  c c | c c c  }
    \Xhline{2\arrayrulewidth}
    
    \multirow{2}{*}{\textbf{Method}} & \multicolumn{7}{c|}{ScanNet Test-Set (m)} & \multicolumn{3}{c}{DTU Test-Set (mm)} \\
    \cline{2-11} 
	& abs-rel & abs & sq-rel & rmse & rmse-log & $\delta <$ 1.25 & $\delta <$ 1.25$^2$ & abs-rel & abs & rmse \\
	\hline
	MVDepth \cite{MVDepthNet2018} & 0.1167 & 0.2301 & 0.0596 & 0.3236 & 0.1610 & 0.8453 & 0.9639 & - & - & -   \\
	MVDepth-FT & 0.1116 & 0.2087 &0.0763 & 0.3143 & 0.1500 & 0.8804 & 0.9734 & - & - & -   \\
	DPSNet \cite{im2019dpsnet} &  0.1200 & 0.2104 & 0.0688 & 0.3139 & 0.1604  & 0.8640 & 0.9612 & - & - & -  \\
	DPSNet-FT & 0.0986 & 0.1998 & 0.0459 & 0.2840 & 0.1348 & 0.8880 & 0.9785  & - & - & -   \\
	DELTAS \cite{deltas-sinha2020} & 0.0915 & 0.1710 & 0.0327 & 0.2390 & 0.1226 & 0.9147 & \textbf{0.9872} & - & - & -  \\
	
	NRGBD \cite{liu2019neural} & 0.1013 & 0.1657 & 0.0502 & 0.2500 &0.1315 & 0.9160 & 0.9790  & - & - & -   \\
	ESTD \cite{Long_2021_CVPR_EstDepth} & \underline{0.0812} & \underline{0.1505} & \underline{0.0298} & \underline{0.2199} & \underline{0.1104} & \underline{0.9313} & \underline{0.9871} &  - & - & -   \\
	\hline
       MVSNet \cite{yao2018mvsnet} &  0.1032 & 0.18645 & 0.0465 & 0.2743 & 0.1385 & 0.8935 & 0.9775
       & 0.0143  & 10.7235 & 25.3989  \\
	PairNet \cite{deepvideomvs2021} &  0.0895 & 0.1709 & 0.0615 &	0.2734 & 0.1208 & 0.9172 &	0.9804  & 0.0129 & 9.4428 & \underline{21.4650}  \\
	IterMVS \cite{itermvs_wang2021} & 0.0991 & 0.1818 &	 0.0518 &	0.2733	& 0.1368 & 0.8995 & 0.9741 & 0.0146 & 10.6225 & 28.7009 \\
	\hline 
	Ours(base)  & 0.0885	& 0.1605 & 0.0380 & 0.2347 & 0.1183 & 0.9211 & 0.9810	
    & \underline{0.0116} & \underline{8.2887} & 21.5806  \\
	
 
	Ours(+pose,atten) & \textbf{0.0734} & \textbf{0.1381}	& \textbf{0.0281} & \textbf{0.2080} & \textbf{0.1030} & \textbf{0.9395}	& {0.9862} & 
 \textbf{0.0092} & \textbf{6.7771} &  \textbf{18.5953} \\	
	
	\Xhline{2\arrayrulewidth}
	\end{tabularx}}

\caption{ Quantitative evaluation results on the test set of ScanNet \cite{dai2017scannet} and the test set of DTU \cite{dtu_jensen2014large}. Error metrics (lower is better) are abs-rel, abs, sq-rel, rmse, rmse-log, while accuracy (higher is better) metrics are $\delta<$ 1.25/1.25$^2$/1.25$^3$. Here \textit{-FT} denotes finetuned on ScanNet. \textit{Bold} is the best score, and \textit{underline} indicates the second best one.
} 
\label{tab:main-scan-dtu}
\end{table*}

\begin{figure*}[t]
    \centering
    \renewcommand{\tabcolsep}{0.2pt}
    \scriptsize
    \begin{tabular}{ccccc}
    
        \includegraphics[width=0.2\textwidth]{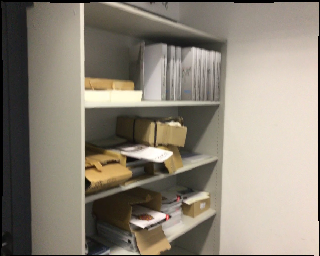} &  \includegraphics[width=0.2\textwidth]{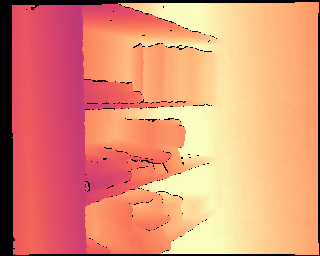} & 
        \begin{overpic}[width=0.2\textwidth]{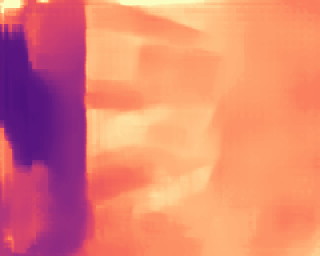}
        \put (5,65) {$\displaystyle\textcolor{white}{\textbf{0.1471}}$}
        \put(56,40){{\color{red} \circle{40}}}
        \end{overpic} &
      \begin{overpic}[width=0.2\textwidth]{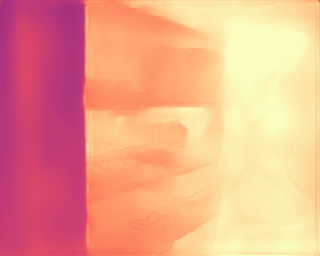}
      \put (5,65) {$\displaystyle\textcolor{white}{\textbf{0.0997}}$}
      \put(56,40){{\color{red} \circle{40}}}
      \end{overpic} &
      \begin{overpic}[width=0.2\textwidth]{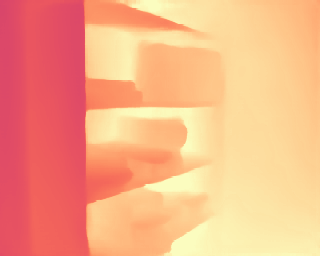} 
      \put (5,65){$\displaystyle\textcolor{white}{\textbf{0.0833}}$}
      \put(56,40){{\color{green} \circle{40}}}
      \end{overpic} \\

        \includegraphics[width=0.2\textwidth]{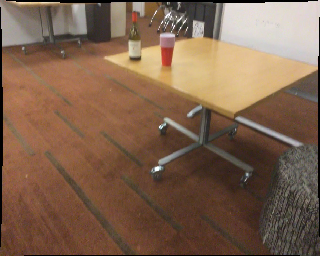} &  \includegraphics[width=0.2\textwidth]{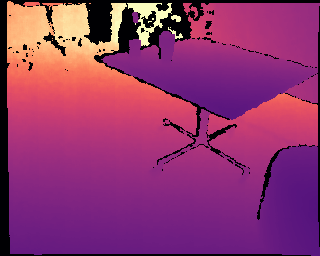} & 
        \begin{overpic}[width=0.2\textwidth]{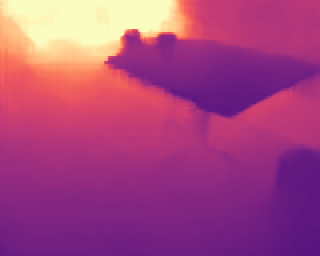}
        \put (72,65){$\displaystyle\textcolor{white}{\textbf{0.0592}}$}
        \put(46,70){{\color{red} \circle{20}}}
        \put(65,35){{\color{red} \circle{30}}}
        \end{overpic} &
      \begin{overpic}[width=0.2\textwidth]{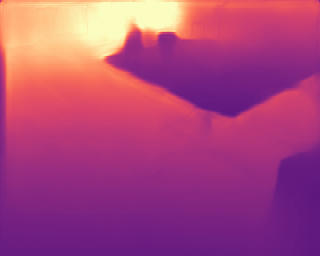}
      \put (72,65){$\displaystyle\textcolor{white}{\textbf{0.0573}}$}
      \put(46,70){{\color{red} \circle{20}}}
      \put(65,35){{\color{red} \circle{30}}}
      \end{overpic} &
      \begin{overpic}[width=0.2\textwidth]{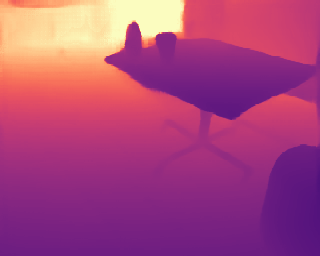} 
      \put (72,65){$\displaystyle\textcolor{white}{\textbf{0.0528}}$}
      \put(46,70){{\color{green} \circle{20}}}
      \put(65,35){{\color{green} \circle{30}}}
      \end{overpic} \\

      \includegraphics[width=0.2\textwidth]{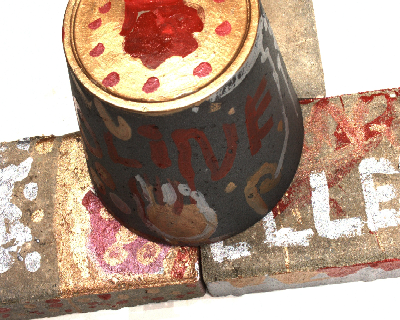} &  \includegraphics[width=0.2\textwidth]{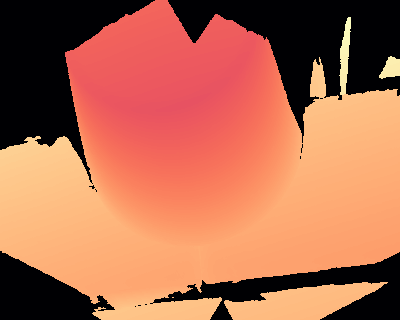} & 
 \begin{overpic}[width=0.2\textwidth]{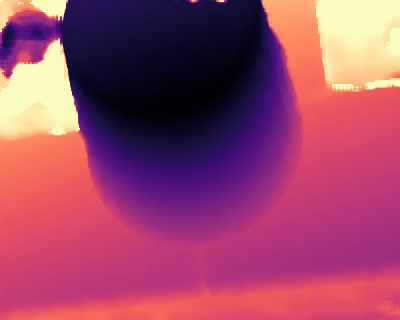}
 \put (40,65){$\displaystyle\textcolor{white}{\textbf{0.0313}}$}
        \put(85,65){{\color{red} \circle{30}}}
        \put(15,65){{\color{red} \circle{30}}}
        \end{overpic}
         & 

        \begin{overpic}[width=0.2\textwidth]{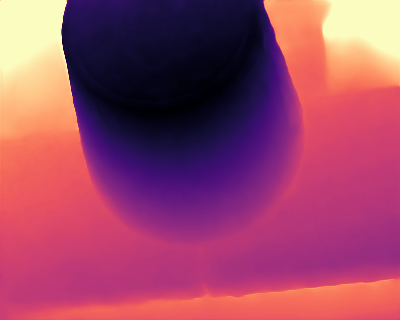}
        \put (40,65){$\displaystyle\textcolor{white}{\textbf{0.0107}}$}
        \put(85,65){{\color{red} \circle{30}}}
        \put(15,65){{\color{red} \circle{30}}}
        \end{overpic} 
        &

      \begin{overpic}[width=0.2\textwidth]{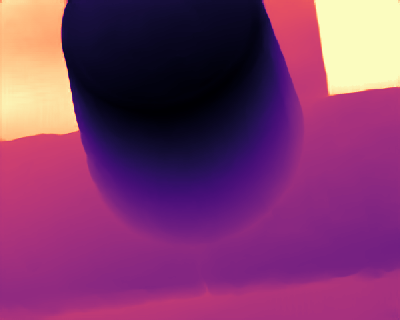}
      \put (40,65){$\displaystyle\textcolor{white}{\textbf{ 0.0042}}$}
        \put(85,65){{\color{green} \circle{30}}}
        \put(15,65){{\color{green} \circle{30}}}
        \end{overpic} \\

        \includegraphics[width=0.2\textwidth]{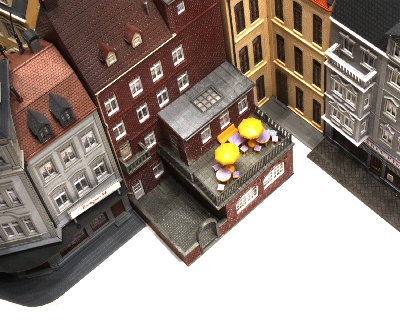} &  \includegraphics[width=0.2\textwidth]{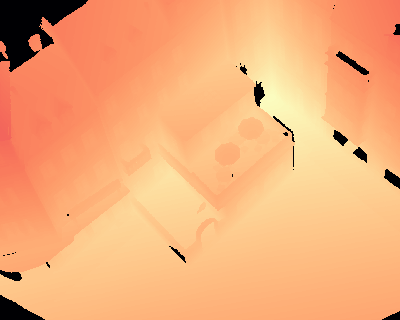} & 

\begin{overpic}[width=0.2\textwidth]{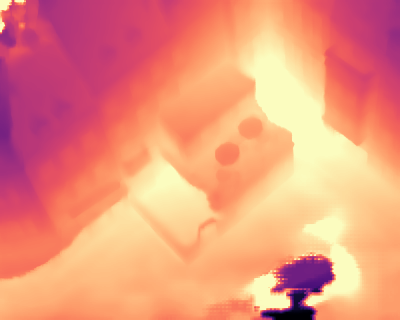}
\put (72,65){$\displaystyle\textcolor{white}{\textbf{0.0093}}$}
        \put(75,15){{\color{red} \circle{30}}}
        \end{overpic}
         & 

        \begin{overpic}[width=0.2\textwidth]{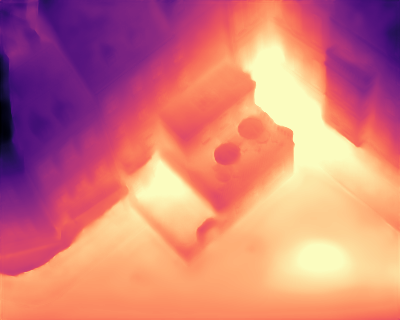}
        \put (72,65){$\displaystyle\textcolor{white}{\textbf{0.0075}}$}
        \put(75,15){{\color{red} \circle{30}}}
        \end{overpic} 
        &

      \begin{overpic}[width=0.2\textwidth]{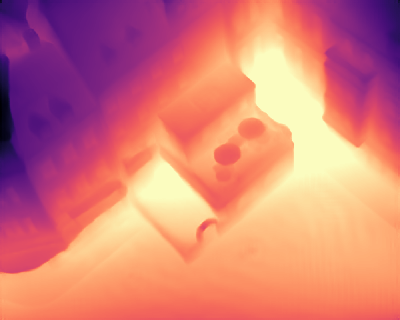}
      \put (72,65){$\displaystyle\textcolor{white}{\textbf{0.0049}}$}
        \put(75,15){{\color{green} \circle{30}}}
        \end{overpic} \\
      
        \normalsize Ref. image & \normalsize GT depth & \normalsize IterMVS & \normalsize PairNet & \normalsize Ours \\
    \end{tabular}
    \vspace{-2pt}
    \caption{Qualitative results on ScanNet \cite{dai2017scannet} (top two rows) and DTU \cite{dtu_jensen2014large} test set. Left two columns show reference image and ground truth depth, and other columns are the estimated depth by baseline IterMVS \cite{itermvs_wang2021}, PairNet \cite{deepvideomvs2021} and ours (the full version), respectively. Our method outperform the baselines on thin structures, small objects and boundaries, as highlighted in {\color{green} green} for ours and in {\color{red} red} for the baselines. The abs-err errors (in meters) are imposed on the depth maps for comparison.
    }
    \label{fig:qualitatives_scannet_dtu}
\end{figure*}
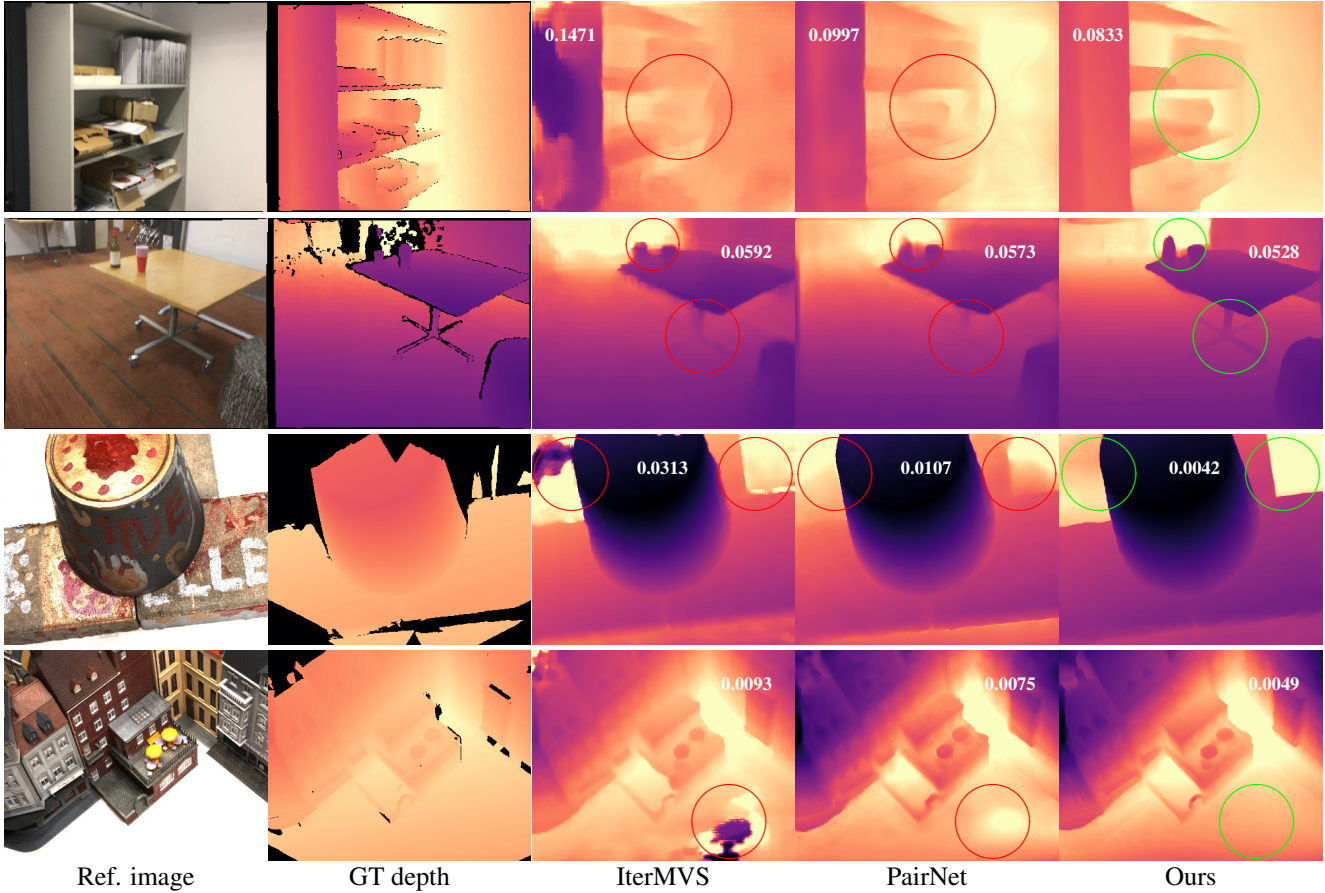

 \noindent \textbf{Quantitative Evaluation.} \,\, 
 We evaluate the depth maps using the standard metrics in \cite{eigen2014depth}, including mean absolute relative error ({\myqcrfont abs-rel}), mean absolute error ({\myqcrfont abs}), squared relative error ({\myqcrfont sq-rel}), root mean square error in linear scale ({\myqcrfont rmse}) and log scale ({\myqcrfont rmse-log}), and inlier ratios under thresholds of $\sigma <$1.25/1.25$^2$/1.25$^3$. Tab. \ref{tab:main-scan-dtu} shows the results on the ScanNet benchmark of our methods and several state-of-the-art MVS methods. We compare two variants of our models: i) \textit{base}: a base version with our recurrent indexing cost volume, and ii) \textit{+pose,atten}: a full version with residual pose and the asymmetric employment of transformer self-attention. Our full model achieves the best performance in most metrics except the inlier ratio under {\myqcrfont $\sigma<$1.25$^2$}, outperformed by DELTAS \cite{deltas-sinha2020} and ESTD \cite{Long_2021_CVPR_EstDepth}. But the inlier ratio is not as essential as {\myqcrfont abs-rel} and {\myqcrfont abs}. \textbf{Zero-shot Generalization.} \,\, 
 We evaluate the generalization performance of our method \ourname{} from ScanNet to other indoor datasets without any fine-tuning. Tab. \ref{tab:scan-2-others} shows the results of the methods trained on ScanNet \cite{dai2017scannet} and directly tested on 7-scenes \cite{7scenes-glocker2013real-time} and RGB-D Scenes V2 \cite{lai2014unsupervised}. Our models outperform the baselines, and our two variants all have strong generalization performance. \textbf{Qualitative Results.} \,\, Fig. 
 \ref{fig:qualitatives_scannet_dtu} demonstrates the qualitative results of our method vs. baselines IterMVS \cite{itermvs_wang2021} and PairNet \cite{deepvideomvs2021} on the test set of ScanNet \cite{dai2017scannet} and DTU \cite{dtu_jensen2014large}. Our method can make more accurate and sharp depth predictions, especially for regions near boundaries and edges. For both near and far objects, our method outperforms the baselines.
 

\begin{table*}[!tb]
\small
\vspace{1pt}
\centering
\scalebox{0.94}{
    \begin{tabularx}{0.95\linewidth}{ l | c c c c c | c c c c c}
    \Xhline{2\arrayrulewidth}
    {\textbf{ScanNet} $\Rightarrow$ } &\multicolumn{5}{c|}{7-Scenes} & 
    \multicolumn{5}{c}{RGB-D Scenes V2} \\
    \cline{2-11} 
	\textbf{Others} & abs-rel & abs 
	& sq-rel & rmse  & $\delta <$ 1.25 & abs-rel & abs 
	& sq-rel & rmse  & $\delta <$ 1.25 \\
	\hline
	NRGBD \cite{liu2019neural} & 0.2334 & 0.4060 & 0.2163 & 0.5358 & 0.6803 & - & - & - & - & - \\
	ESTD \cite{Long_2021_CVPR_EstDepth}  & 0.1465 & 0.2528 & 0.0729 & 0.3382 & 0.8036 & - & - & - & - & - \\
	\hline
	PairNet \cite{deepvideomvs2021} 
	& 0.1157 & 0.2086 & 0.0677 & 0.2926 & \underline{0.8768}  
	& 0.0995 & 0.1382 &  0.0279 & 0.1971 & 0.9393 \\
	IterMVS \cite{itermvs_wang2021}
	& 0.1336 & 0.2363 & 0.1033 & 0.3425 & 0.8518 
	& \underline{0.0811} & \underline{0.1245} & 0.0340 & 0.2133 & \underline{0.9496} \\
	
	\hline
    Ours(base) 
    & \underline{0.1148} & \underline{0.1999} &  \underline{0.0552} & \underline{0.2857} & 0.8726 
    & 0.0967 & 0.1336 & \underline{0.0246} & \underline{0.1836} & 0.9427   \\
    
    Ours(+pose,atten) 
    & \textbf{0.1000} & \textbf{0.1781} & \textbf{0.0473} & \textbf{0.2664} & \textbf{0.8967} 
   
    & \textbf{0.0803} & \textbf{0.1168} & \textbf{0.0200} & \textbf{0.1703} & \textbf{0.9632}  \\
    
	\Xhline{2\arrayrulewidth}
	\end{tabularx}}
\vspace{1pt}
\caption{ Zero-shot generalization from ScanNet \cite{dai2017scannet} to 7-scenes \cite{7scenes-glocker2013real-time} and RGB-D Scenes V2 \cite{lai2014unsupervised}. Our methods achieve better generalization. We sample the sequences every 10 frames, and each sample has 5 frames for multi-view stereo depth prediction. 
} 
\label{tab:scan-2-others}
\end{table*}

 \begin{table*}[!htb]
\begin{tabular}{c c c }
    \begin{minipage}{.4\linewidth}
      \scalebox{0.8}{
      
        \begin{tabular}{l | c c c c}
        \Xhline{2\arrayrulewidth}
        \multirow{2}{*}{\textbf{Variants}} & \multicolumn{4}{c}{ScanNet Test-Set (m)} \\
        \cline{2-5}
          & abs-rel & abs & rmse & $\delta <$ 1.25  \\
	\hline

	Ours(base)  & 0.0885	& 0.1605 & 0.2347 & 0.9211 \\
	Ours(+pose)  & 0.0827 & 0.1523	&  0.2253 &  0.9277 \\
 
	Ours(+pose,atten) & 
 \textbf{0.0734} & \textbf{0.1381}	& \textbf{0.2080} & \textbf{0.9395}	\\
 
	\Xhline{2\arrayrulewidth}
	
    \end{tabular}
    }
    \caption*{\small {(a) We compare three variants of our models.}}
    \end{minipage}  &  &
    
    \begin{minipage}{0.5\linewidth}
      \scalebox{0.76}{
      
        \begin{tabular}{l | c c c c}
        \Xhline{2\arrayrulewidth}
        \multirow{2}{*}{\textbf{Attention}} & \multicolumn{4}{c}{ScanNet/DTU Test-Set (m)} \\
        \cline{2-5}
          & abs-rel & abs & rmse & $\delta <$ 1.25  \\
	\hline
	Asym atten (ours) & 
 \textbf{0.0734} & \textbf{0.1381}	& \textbf{0.2080} & \textbf{0.9395}	\\
        Sym. atten & 0.0761 & 0.1496 & 0.2253	& 0.9333	\\
        \hline
        \multirow{2}{*}{MVSNet \cite{yao2018mvsnet}}
         & 0.1032 & 0.1865  & 0.2743 & 0.8935 \\
         & (0.0143) & (10.7235) & (25.3989) & (0.8936) \\
         \cline{2-5}
         \multirow{2}{*}{MVSNet(+atten)}
         & 0.1018  & 0.1853 & 0.2734 & 0.8957  \\
         & (0.0123) & (9.1150) & (22.3525) & (0.9909) \\
	\Xhline{2\arrayrulewidth}
	
    \end{tabular}
    }
    \caption*{\small {(b) Asymmetric attention and MVSNet backbone. DTU results are in parentheses.}}
    \end{minipage}

\end{tabular}
\vspace{-6pt}
    \caption{ Ablation study for our proposed modules and a different backbone architecture MVSNet \cite{yao2018mvsnet}.
} 
\label{tab:ablation-task2}
\end{table*}

 
 \subsection{Ablation Study}
 
 \begin{table*}[!htb]
\begin{tabular}{c c c c c }
    \begin{minipage}{.26\linewidth}
      \scalebox{0.8}{
        \begin{tabular}{l | c c c}
        \Xhline{2\arrayrulewidth}
         Itr. T & abs-rel & abs & $\delta <$ 1.25  \\
	\hline
	16  & 0.1413 & 0.0760 & 0.9364 \\
    24 & 0.1400 & 0.0752 & 0.9375 \\
    48 & 0.1392 & 0.0747 & 0.9382 \\
    64 & 0.1392 & 0.0746 & 0.9384 \\
    96 & \textbf{0.1392} & \textbf{0.0745} & \textbf{0.9385} \\
    128 & \underline{0.1394} & \underline{0.0745} & \underline{0.9385} \\
	\Xhline{2\arrayrulewidth}
	
    \end{tabular}
    }
    \caption*{\small {(a) GRU iterations}}
    \end{minipage}  &  &
    
    \begin{minipage}{0.3\linewidth}
      \scalebox{0.8}{
        \begin{tabular}{l | c c c }
        \Xhline{2\arrayrulewidth}
        
        \hline
        View No. & abs-rel & abs & $\delta <$ 1.25  \\
	\hline
	3 (base) & 0.1204 & 0.2121 & 0.8603 \\
    5 (base) & 0.1148 & 0.1999 & 0.8726\\
    \hline
    3 (+pose) & 0.1162 & 0.2061 & 0.8711 \\
    5 (+pose) & 0.1096 & 0.1930 & 0.8840 \\
    \hline
    3 (+pose,atten) &  \underline{0.1084} & \underline{0.1923} &\underline{0.8833} \\
    5 (+pose,atten) & \textbf{0.1000} & \textbf{0.1781} & \textbf{0.8967} \\
	\Xhline{2\arrayrulewidth}
	
    \end{tabular}
    }
    \caption*{\small {(b) View numbers}}
    \end{minipage}  &  &
    
    \begin{minipage}{.32\linewidth}
      \scalebox{0.80}{
        \begin{tabular}{l | c c c}
        \Xhline{2\arrayrulewidth}
        Sampling & abs-rel & abs & $\delta <$ 1.25  \\
	\hline
	s10 (base) & 0.0885 & 0.1605 & 0.9211 \\
	key (base) & 0.0838 & 0.1598 & 0.9277 \\
    
    \hline
    s10 (+pose) & 0.0827 & 0.1523 & 0.9277 \\
    key (+pose) & 0.0789 & 0.1531 & 0.9339 \\
    
    \hline
    s10 (+pose,atten) & \underline{0.0747} & \underline{0.1392} & \underline{0.9382} \\
    key (+pose,atten) &  \textbf{0.0697} & \textbf{0.1348} & \textbf{0.9472} \\
	\Xhline{2\arrayrulewidth}
	
    \end{tabular}
    }
    \caption*{\small {(c) Frame sampling}}
    \end{minipage} 
    
\end{tabular}
\vspace{-6pt}
    \caption{ Ablation study of design choices. \textit{Bold} is the best, and \textit{underline} indicates the second best.
} 
\label{tab:ablation-scan}
\end{table*}

 \begin{table}[!tb]
\small
\vspace{-5pt}
\centering
\scalebox{0.76}{
    \begin{tabularx}{1.0\linewidth}{ l | c c  c c}
    \Xhline{2\arrayrulewidth}
    View No. on DTU & abs-rel ($\downarrow$) & abs (mm) ($\downarrow$) & rmse ($\downarrow$) \\
    \hline
	3 (1 ref + 2 source) & 0.0149 & 11.0689 &  24.8831 \\
    
    
    5 (1 ref + 4 source) & 0.0119 & 8.8419 & 21.4327 \\
	\Xhline{2\arrayrulewidth}
	\end{tabularx}
}
\vspace{-8pt}
\caption{ \small {3-view vs. 5-view training and testing on DTU \cite{dtu_jensen2014large}.}} 
\label{tab:rebu-view-num}
\end{table}

\begin{table}[!tb]
\small
\vspace{-4pt}
\centering
\scalebox{0.76}{
    \begin{tabularx}{1.1\linewidth}{ l | c c  c c }
    \Xhline{2\arrayrulewidth}
    Methods & Time(fps) & Mem.(MB)  & Param.(M) & abs-rel ($\downarrow$) \\
    \hline
	Ours(T=8) &  6.98 &  4297 &  27.6 & 0.0760\\
	Ours(T=12) & 5.91 &  4297 &  27.6 & 0.0752 \\
	Ours(T=24) & 3.77 &  4297 &  27.6 &  0.0734 \\
	\hline
    IterMVS \cite{itermvs_wang2021} 
    & 22.61 & 2171 & 0.34 & 0.0991 \\
    ESTD \cite{Long_2021_CVPR_EstDepth} 
    & 14.08 & 1799 & 36.2 & 0.0812 \\
	\Xhline{2\arrayrulewidth}
	\end{tabularx}
}
\vspace{-8pt}
\caption{ \small{Comparison of run time, memory consumption, and accuracy on ScanNet \cite{dai2017scannet} test set with frame dimension $320 \times 256$.}} 
\label{tab:rebu-runtime}
\end{table}

\vspace{2pt}
 \noindent \textbf{Efficacy of Proposed Modules:} Our design is verified by ablating the modules to three variants, as shown in Tab.~\ref{tab:ablation-task2}-(a). The \textit{base} version itself can achieve competitive performance on ScanNet and better generalization, verifying the efficacy of our novel design - cost volume recurrent indexing via index field. Further, the performance can be consistently boosted when the residual pose net (i.e., variant \textit{+pose}) and transformer self-attention are added (i.e., variant \textit{+pose,atten}). Therefore, each of the proposed modules can consistently help with accurate depth estimation. Tab.~\ref{tab:ablation-task2}-(b) shows the benefit of using asymmetric attention over symmetric attention. We also see improvements when applying our asymmetric attention to the MVSNet \cite{yao2018mvsnet} backbone on ScanNet and DTU test sets (see the results in parentheses). \textbf{Number of GRU Iterations and Convergence:} Tab.~\ref{tab:ablation-scan}-(a) shows the ablation study on different number of GRU iterations $T$. The results are obtained by running our model (the full version) on the ScanNet test set, with $T=$16, 24, 48, 64, 96, and 128. Running more iterations boosts our depth prediction, but after $T \geq 96$, the gain is marginal. \textbf{View Number:} We compare 3-view (\ie, 1 reference + 2 source images) and 5-view (\ie, 1 reference + 4 source images). Tab.~\ref{tab:ablation-scan}-(b) shows that the more frames are used for matching, the better the depth will be. The results are obtained for the zero-shot generalization from ScanNet to 7-Scenes. Note that our full model (\textit{+pose,atten}) with 3-view input outperforms the other two variants with 5-view input, showing the asymmetrical employment of the transformer self-attention can boost the prediction due to the mining of more global information. \textit{3-view vs 5-view on DTU}: Tab.~\ref{tab:rebu-view-num} shows that our model, \ie, Ours(+pose,atten) is trained/tested on DTU dataset with 3-view and 5-view input collections, respectively. We use the same training scheduling for a fair comparison. The 5-view result is worse than the last row in Tab.~\ref{tab:main-scan-dtu} due to the lack of pretraining on ScanNet. \textbf{Frame Sampling:} We compare the simple view selection strategy (i.e., sampling by every 10 frames) with the heuristics \cite{deepvideomvs2021}. Tab.~\ref{tab:ablation-scan}-(c) shows that ours can be improved when the selected views have more overlapping and the baselines are suitable. Our(+pose,atten) even with simple strategy outperforms other variants with heuristic sampling, and so are our(+pose) vs our(base). \textbf{Runtime Overhead:} Tab.~\ref{tab:rebu-runtime} shows the run-time and memory consumption when processing $320 \times 256$ frames from the ScanNet test set. Ours (T=8/12/24) means 8, 12, and 24 GRU iterations.

\section{Conclusions}\label{sec:conclusions}
We have proposed \ourname, a novel learning-based MVS method. Our approach utilizes a convolutional GRU to iteratively optimize the index fields, which are used to access the cost volume and regress the depth. The cost volume is further improved through the application of a transformer block to the reference image and a residual pose network to correct the relative poses. Extensive experiments on ScanNet~\cite{dai2017scannet}, DTU \cite{dtu_jensen2014large}, 7-Scenes~\cite{7scenes-glocker2013real-time}, and RGB-D Scenes V2~\cite{lai2014unsupervised} have demonstrated the superior accuracy and cross-dataset generalizability of our method. Due to the plane-sweeping 3D cost volume and transformer self-attention, our method requires large memory consumption for high-resolution images. Moreover, the inference time is not as fast as other lightweight convolutional counterparts, due to the iterative update paradigm in our approach. In future work, we plan to leverage temporal information to further enhance depth estimation from posed-video streams.


{\small
\bibliographystyle{ieee_fullname}
\bibliography{my_ref}
}

\appendix

In this supplementary material, we show more details about datasets, network architectures and hyperparameters, ablation studies, and additional qualitative results.

\section{Datasets}
Our experiments use four indoor-scene datasets, which have RGB-D video frames with ground truth depths and known camera poses. ScanNet \cite{dai2017scannet} and DTU \cite{dtu_jensen2014large} are used in training and testing, and 7scenes \cite{7scenes-glocker2013real-time} and RGB-D Scenes V2 \cite{lai2014unsupervised} are evaluated for zero-shot generalization. 

\vspace{5pt}
\noindent \textbf{ScanNet}. \,\, Our network is trained from scratch on ScanNet \cite{dai2017scannet} using the official training split. Following the frame selection heuristic in \cite{deepvideomvs2021}, considering appropriate view frustum overlap and sufficient baselines, we sample 279,494 training samples and 20,000 validation ones. Each sample contains 3 frames, with one as a reference frame and the others as source frames. For testing, we use ScanNet's official test split (with 100 sequences from scene707 to scene806) and sample every 10 frames following \cite{Long_2021_CVPR_EstDepth}, resulting in 20,668 samples for quantitative evaluation. ScanNet has images in 640$\times$480 resolution. In training, they are resized to 256$\times$256 with cropping following \cite{deepvideomvs2021}. For inference, the input images are resized to 320$\times$256 without cropping. The predicted depth maps are upsampled with nearest neighbor interpolation to the original resolution 640$\times$480 before calculating the quantitative metrics. 

\vspace{5pt}
\noindent \textbf{DTU}.\,\, DTU~\cite{dtu_jensen2014large} is a smaller dataset compared with ScanNet, but with accurate ground truth depth and pose obtained by a structured light scanner. Following \cite{patchmatchnet_wang2020,yao2018mvsnet,yao2019recurrent}, the depth range for sampling depth hypotheses is set to $d_{min} = 0.425$ and $d_{max} = 0.935$ meters. Based on the view selection and robust training strategy in \cite{patchmatchnet_wang2020,itermvs_wang2021}, we sample 27,097 training samples, 6,174 validation ones, and 1,078 ones for evaluation. Each sample has 5 frames. Input image size is 512$\times$256 in network training, and 640$\times$512 for inference and upsampled with nearest neighbor interpolation to the original size 1600$\times$ 1152 for evaluation. To coordinate with ScanNet \cite{dai2017scannet} and 7scenes \cite{7scenes-glocker2013real-time}, we use the same depth evaluation metrics proposed in \cite{eigen2014depth}. 

\vspace{5pt}
\noindent \textbf{7-Scenes}.\,\, We select 13 sequences from 7-Scenes for zero-shot generalization. The valid depth range is set the same as that on ScanNet. We generate a test set with 1,610 samples (each with 5 frames, at 640$\times$480 resolution) by sampling the sequences every 10 frames. 

\vspace{5pt}
\noindent \textbf{RGB-D Scenes V2}.\,\, It contains indoor scenes, including chair, sofa, table, bowls, caps, cereal boxes, coffee mugs, and soda cans, etc. We select 8 sequences for testing. Similarly, we sample the video sequence every 10 frames to generate 610 testing samples (each with 5 frames). 

\section{Experimental Setup}
 
 \noindent \textbf{Implementation Details:} \,\, Our model is implemented using PyTorch \cite{PyTorch-NEURIPS2019_9015}, and trained end-to-end with a mini-batch size of 8 per NVIDIA RTX A6000 GPU. During training, we use the AdamW optimizer and clip gradients to the range of $[-1, 1]$. When generating the cost volume by plane-sweep stereo, we set the plane hypotheses number as $M_0$=64. When predicting the final depth using the index field, we set the plane hypotheses number as $M_1$=256. The same hyperparameters as in \cite{raft-stereo-lipson2021} are adopted for the context network and 3-level GRU architecture.
 
 \vspace{5pt}
 \noindent \textbf{Training Schedule:} \,\, Our network is trained for 20 epochs, with an initial learning rate of 1e-4 and decayed by half at epoch $4^{th}$ and $8^{th}$, respectively. For a fair comparison, we also train the baselines PairNet \cite{deepvideomvs2021} and IterMVS \cite{itermvs_wang2021} on the same training samples of ScanNet for 20 epochs, using the official codes. 
 For the baseline PairNet we follow the suggested learning rate scheduler, and for the baseline Iter-MVS, we use a learning rate of 1e-4, which is decayed by half at epoch $4^{th}$ and $8^{th}$.
 
\section{Our Modules Improve Existing Backbones}

Our proposed residual pose module and asymmetric attention module can help improve existing state-of-the-art methods. Here we take two baselines - IterMVS \cite{itermvs_wang2021} and MVSNet \cite{yao2018mvsnet} as the backbone. \cref{tab:supp-ours-to-mvsnet-backbone}-(a) shows the improved accuracy on the ScanNet test set \cite{dai2017scannet} due to incorporating our residual pose module (\ie, \textit{+pose}) and our asymmetric attention module (\ie, \textit{+atten}). Results in parenthesis and highlighted
by gray, denote the residual pose is only used for network training but not for inference \footnote{only the ground truth pose is used for feature warping and cost volume construction.}. Note that they are listed for reference only, and are not used for comparison with the numbers on other rows. We can see our \textit{+pose} and \textit{+atten} can always boost the baseline backbones on the ScanNet test set. \cref{tab:supp-ours-to-mvsnet-backbone}-(b) shows the evaluation on DTU test set \cite{dtu_jensen2014large}. Our \textit{+atten} always helps improve the baselines. Our \textit{+pose} can boost the baseline IterMVS \cite{itermvs_wang2021}, but achieves no obvious improvement on baseline MVSNet \cite{yao2018mvsnet}, probably because the ground truth poses are accurate enough, and the features are concatenated when constructing the cost volume, which is different from the dot production of features in ours and baseline IterMVS.

\begin{table*}[!tb]
\small
\vspace{1pt}
\centering
\scalebox{1.1}{
    \begin{tabularx}{0.88\linewidth}{ l | c c  c c  c | c c c}
    \Xhline{2\arrayrulewidth}
    
    \multirow{2}{*}{\textbf{Method}} & \multicolumn{8}{c}{ScanNet Test-Set} \\
    \cline{2-9} 
	& abs-rel ($\downarrow$) & abs($\downarrow$)  & sq-rel($\downarrow$) & rmse($\downarrow$) & rmse-log($\downarrow$) & \multicolumn{3}{c}{$\delta <$ 1.25/1.25$^2$/1.25$^3$ ($\uparrow$)} \\
	\hline
       MVSNet \cite{yao2018mvsnet} &  0.1032 & 0.1865 & 0.0465 & 0.2743 & 0.1385 & 
       0.8935 & 0.9775  & \underline{0.9942} \\
        MVSNet(+pose) & \textbf{0.0937} & \textbf{0.1714}	& \textbf{0.0401} & \textbf{0.2565} & \textbf{0.1300}	& 
        \textbf{0.9072} & \textbf{0.9803} & \textbf{0.9947} \\
         & \cellcolor{mygray}(0.0955) & \cellcolor{mygray}(0.1766) & \cellcolor{mygray}(0.0431) & \cellcolor{mygray}(0.2654)	& \cellcolor{mygray}(0.1339) & \cellcolor{mygray}(0.9021) &	\cellcolor{mygray}(0.9785) & \cellcolor{mygray}(0.9941) \\
        MVSNet(+atten) & \underline{0.1018} & \underline{0.1853} & \underline{0.0468} & \underline{0.2734} & \underline{0.1377} & 
        \underline{0.8957} & \underline{0.9779} & 0.9941 \\
        \hline
	IterMVS \cite{itermvs_wang2021} & 0.0991 & 0.1818 &	 0.0518 &	0.2733	& 0.1368 & 
    0.8995 & 0.9741 & 0.9915 \\
	
        IterMVS(+pose) & \underline{0.0958} & \underline{0.1813} & \underline{0.0480} & \underline{0.2715} & \underline{0.1343}	& \underline{0.9004} & \underline{0.9758} & \underline{0.9923} \\
         & \cellcolor{mygray}(0.0943) & \cellcolor{mygray}(0.1777) &	\cellcolor{mygray}(0.0472) & \cellcolor{mygray}(0.2687) & \cellcolor{mygray}(0.1336) & \cellcolor{mygray}(0.9037) &	\cellcolor{mygray}(0.9764) & \cellcolor{mygray}(0.9923) \\
        
        IterMVS(+atten) & \textbf{0.0920} & \textbf{0.1741} & \textbf{0.0431} &\textbf{ 0.2620}	& \textbf{0.1298} &
        \textbf{0.9066}	& \textbf{0.9785} & \textbf{0.9936} \\
        
	\Xhline{2\arrayrulewidth}
	\end{tabularx}}

\caption*{ (a) Quantitative results on ScanNet Test Set \cite{dai2017scannet}.}

\vspace{1pt}
\centering
\scalebox{1.1}{
    \begin{tabularx}{0.87\linewidth}{ l | c c  c c  c  | c c c}
    \Xhline{2\arrayrulewidth}
    
    \multirow{2}{*}{\textbf{Method}} & \multicolumn{8}{c}{DTU Test-Set} \\
    \cline{2-9} 
	& abs-rel ($\downarrow$) & abs($\downarrow$)  & sq-rel($\downarrow$) & rmse($\downarrow$) & rmse-log($\downarrow$) & \multicolumn{3}{c}{$\delta <$ 1.25/1.25$^2$/1.25$^3$ ($\uparrow$)} \\
	\hline
       MVSNet \cite{yao2018mvsnet} &  \underline{0.0143} & \underline{10.7235}	& 1.4193 & 25.3989 & 0.0356 & 0.9882 & 0.9984 & 1.0 \\
       
        MVSNet(+pose) & 0.0151 & 11.1539 & \underline{1.2867}	& \underline{24.3420} &	\underline{0.0337}	& \underline{0.9907}	& \textbf{0.9988}	& \underline{1.0} \\
         & \cellcolor{mygray}(0.0129) & \cellcolor{mygray}(9.8094) & \cellcolor{mygray}(1.2638) & \cellcolor{mygray}(23.8917)	& \cellcolor{mygray}(0.0330)	& \cellcolor{mygray}(0.9905)	& \cellcolor{mygray}(0.9987)	& \cellcolor{mygray}(1.0) \\

        MVSNet(+atten) &  \textbf{0.0123} & \textbf{9.1150}	& \textbf{1.1311}	& \textbf{22.3525}	& \textbf{0.0311}	& \textbf{0.9909}	& \underline{0.9986}	& \textbf{1.0} \\
        \hline
        IterMVS \cite{itermvs_wang2021} & 0.0146 & 10.6225 & 2.1377 & 28.7009 & 0.0404 & \underline{0.9832} & 0.9960	& \underline{0.9997} \\

        IterMVS(+pose) & \textbf{0.0129} & \underline{9.9510} & \textbf{1.8261} &	\underline{28.1695} &	\underline{0.0385} &	0.9831	& \textbf{0.9978} & \textbf{0.9999} \\
        & \cellcolor{mygray}(0.0128) & \cellcolor{mygray}(9.8926) &	\cellcolor{mygray}(1.8216)	& \cellcolor{mygray}(28.1242) & \cellcolor{mygray}(0.0384) & \cellcolor{mygray}(0.9832)	& \cellcolor{mygray}(0.9977) & \cellcolor{mygray}(0.9999) \\
        
        IterMVS(+atten) & \underline{0.0130} & \textbf{9.4121} & \underline{1.8775}	& \textbf{25.6287}	& \textbf{0.0357}	& \underline{0.9860} & \underline{0.9969} & 0.9993  \\
        
	\Xhline{2\arrayrulewidth}
	\end{tabularx}}
\caption*{ (b) Quantitative results on DTU Test Set \cite{dtu_jensen2014large}.}

\caption{ Quantitative evaluation results on the test set of ScanNet \cite{dai2017scannet} and DTU \cite{dtu_jensen2014large} for our modules applied to baseline MVSNet \cite{yao2018mvsnet} and IterMVS \cite{itermvs_wang2021}. Error metrics (lower is better) are abs-rel, abs, sq-rel, rmse, rmse-log, while accuracy (higher is better) metrics are $\delta<$ 1.25/1.25$^2$/1.25$^3$. \textit{Bold} is the best score, and \textit{underline} indicates the second best one. The results given in parenthesis and highlighted by gray, denote that the residual pose is only used for network training, but not for inference. They are listed for reference but not for comparison with other rows. 
} 
\label{tab:supp-ours-to-mvsnet-backbone}
\end{table*}

\section{Network Architectures}

\begin{figure*}[htp]
    \centering
    \renewcommand{\tabcolsep}{0.2pt}
    \scriptsize
      \includegraphics[width=0.8\textwidth]{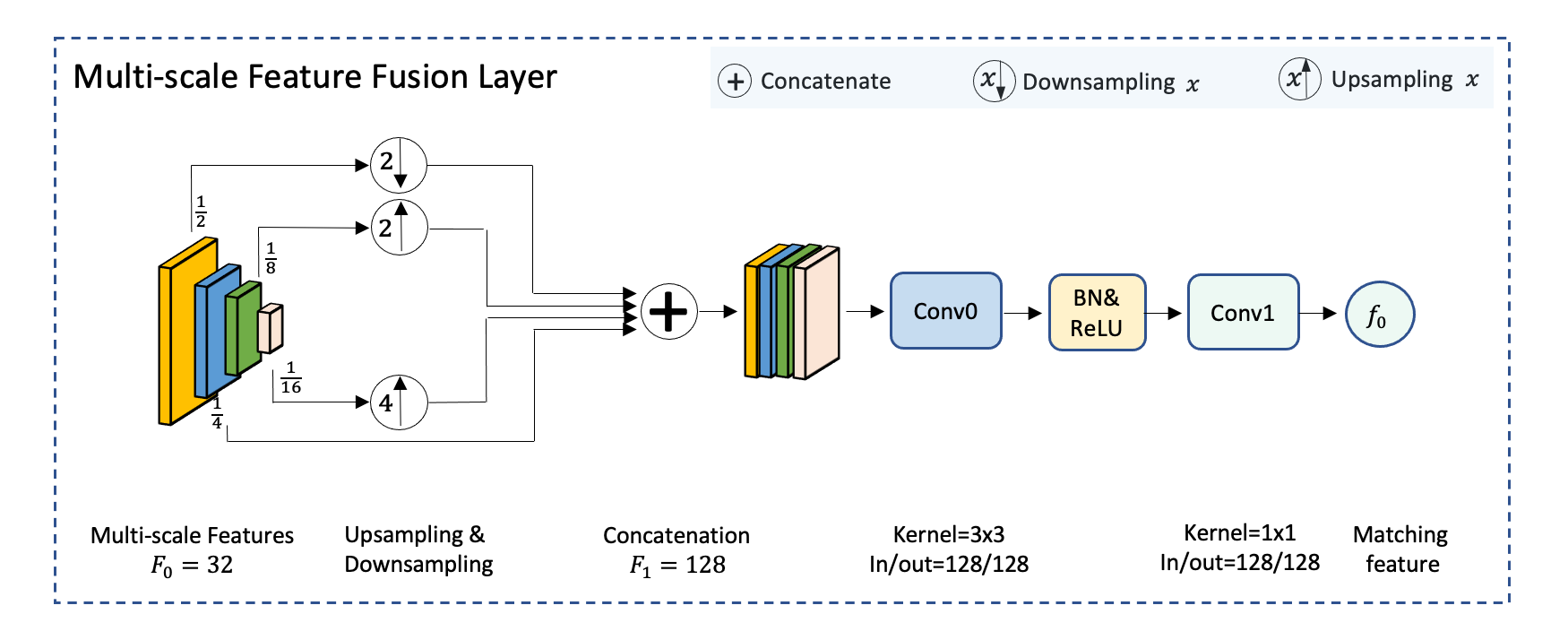}
    \caption{ Multi-scale feature fusion layer.}
    \label{fig:muti-scale-fusion}
\end{figure*}

\noindent \textbf{Multi-scale Feature Fusion Layer.} \,\, The fusion layer $\mathcal{G}$ aggregates multi-scale features 
$f_{i,2} \in \mathbb{R} ^{H/2 \times W/2 \times F_0}$, $f_{i,4} \in \mathbb{R} ^{H/4 \times W/4 \times F_0}$,
$f_{i,8} \in \mathbb{R} ^{H/8 \times W/8 \times F_0}$ and 
$f_{i,16} \in \mathbb{R} ^{H/16 \times W/16 \times F_0}$ into a matching feature $f_i \in \mathbb{R} ^{H/4 \times W/4 \times F_1} $ at 1/4 scale. Here $F_0$=32 and $F_1$=128 for feature channels, $i=0$ for the reference image, and $i=1, \dots, N-1$ for the source images. The architecture is shown in Fig.~\ref{fig:muti-scale-fusion}, including up- and down-sampling, concatenation along the feature channel, a convolution layer {\myqcrfont Conv0} (with kernel size 3$\times$3, in- and out- channels 128/128), batch normalization, ReLU, and another convolution layer {\myqcrfont Conv1} (with kernel size 1$\times$1, in- and out- channels 128/128).

\vspace{5pt}
\noindent \textbf{Context Feature Network C-Net.} \,\, We use the context feature network as in \cite{gma_jiang2021learning,raft-stereo-lipson2021,raft-teed2020}, which consists of several residual blocks. It contains around 4.32M parameters.


\begin{table*}[!htb]
\centering
\scalebox{1.1}{
\begin{tabularx}{0.8\linewidth}{r | c c c c c | r}
\Xhline{2\arrayrulewidth}
    Layers & F-Net &  C-Net & Transformer &  Residual Pose Net & GRUs & Total \\
	\hline
	Parameter (M) & 2.9545 & 4.3212 & 0.3438 & 13.0120 & 6.9501 & 27.5816 \\
	\hline
	Percentage  & 10.70\%  & 15.67\% & 1.25\% & 47.18\% & 25.20\% & 100\% \\
	\Xhline{2\arrayrulewidth}
	
\end{tabularx}
}
\vspace{2pt}
\caption*{ (a) Our model capacity (full version).}

\vspace{2pt}
\centering
\scalebox{1.1}{
\begin{tabularx}{0.8\linewidth}{r | c c c c c | r}
\Xhline{2\arrayrulewidth}
    Layers & F-Net &  C-Net & Transformer &  Residual Pose Net & GRUs & Total \\
	\hline
	Parameter (M) & 2.9545 & 4.3212 & 0.3438 & - & 6.9501 & 14.5696 \\
	\hline
	Percentage  & 20.28\%  & 29.66\% & 2.36\% & - & 47.70\% & 100\% \\
	\Xhline{2\arrayrulewidth}
	
\end{tabularx}
}
\vspace{2pt}
\caption*{ (b) Our model capacity, if without residual pose net.}

    \caption{Our model capacity. Parameter numbers are given in million (M) and the percentage of each module is listed.
} 
\label{tab:supp-model-capacity}
\end{table*}

\vspace{5pt}
\noindent \textbf{Model Capacity.} \,\, As shown in Tab.~\ref{tab:supp-model-capacity}, the total number of parameters in our network is 27.6M, where residual pose network takes up 47.18\%, GRU-based optimizer takes up 25.20\%, and the transformer block takes up 1.25\%. If not considering the residual pose net, our model then has 14.57M parameters, and most of them are assigned to GRU-based updater, and fewer capacities are on feature extractors. This kind of capacity configuration makes our model not specialized to one domain (for feature extraction), and is well generalized to unseen domains due to the learning to optimize anchored at cost volume via the GRU-based optimizer to predict the index fields for iteratively improved matching.

\begin{figure*}[htp]
    \centering
    \renewcommand{\tabcolsep}{0.2pt}
    \scriptsize
    \begin{tabular}{cccc}
      \includegraphics[width=0.22\textwidth]{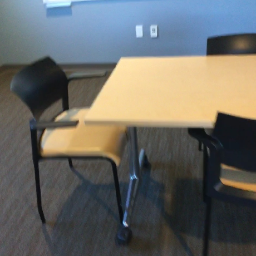} &  \includegraphics[width=0.22\textwidth]{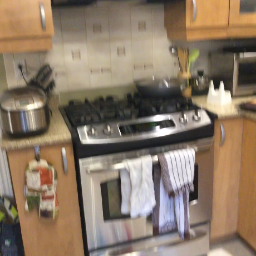} & 
        \includegraphics[width=0.22\textwidth]{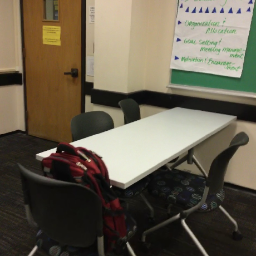} & \includegraphics[width=0.22\textwidth]{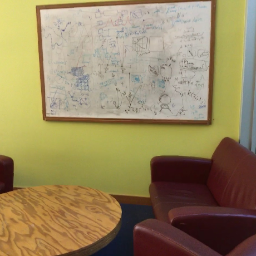} 
        \\
        \normalsize Ref $I_0$ (b0) & \normalsize Ref $I_0$ (b1) & \normalsize Ref $I_0$ (b2) & \normalsize Ref $I_0$ (b3) \\
        
     \includegraphics[width=0.22\textwidth]{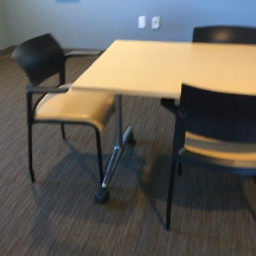} &  \includegraphics[width=0.22\textwidth]{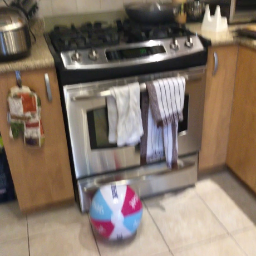} & 
        \includegraphics[width=0.22\textwidth]{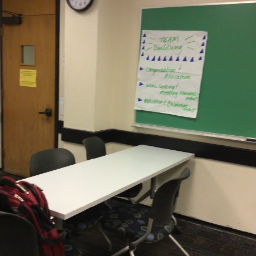} & \includegraphics[width=0.22\textwidth]{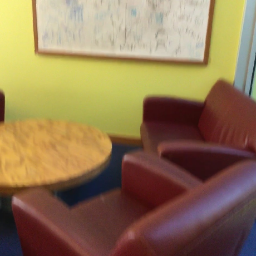} 
        \\
        \normalsize Src $I_1$ (b0) & \normalsize Src $I_1$ (b1) & \normalsize Src $I_1$ (b2) & \normalsize Src $I_1$ (b3) \\
    
    \includegraphics[width=0.22\textwidth]{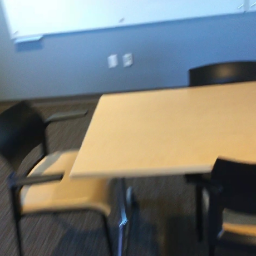} &  \includegraphics[width=0.22\textwidth]{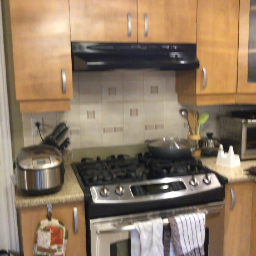} & 
        \includegraphics[width=0.22\textwidth]{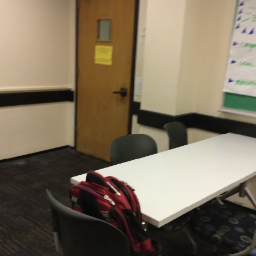} & \includegraphics[width=0.22\textwidth]{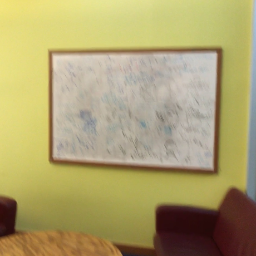} 
        \\
        \normalsize Src $I_2$ (b0) & \normalsize Src $I_2$ (b1) & \normalsize Src $I_2$ (b2) & \normalsize Src $I_2$ (b3) \\
        
    \includegraphics[width=0.22\textwidth]{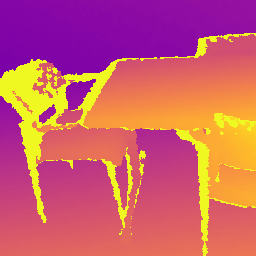} &  \includegraphics[width=0.22\textwidth]{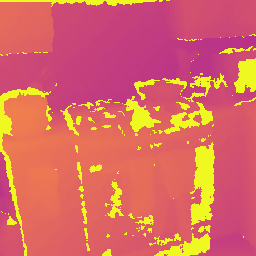} & 
        \includegraphics[width=0.22\textwidth]{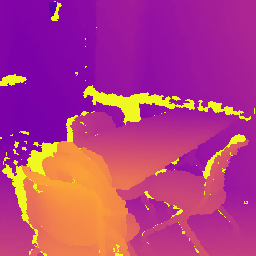} & \includegraphics[width=0.22\textwidth]{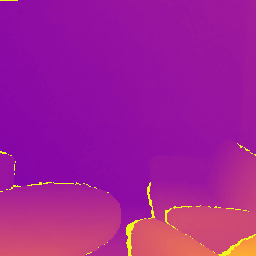} 
        \\
        \normalsize GT Depth (b0) & \normalsize GT Depth (b1) & \normalsize GT Depth (b2) & \normalsize GT Depth (b3) \\
    
    \includegraphics[width=0.22\textwidth]{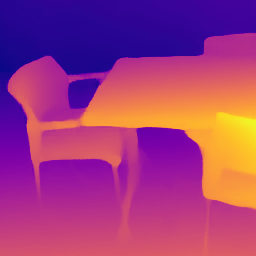} &  \includegraphics[width=0.22\textwidth]{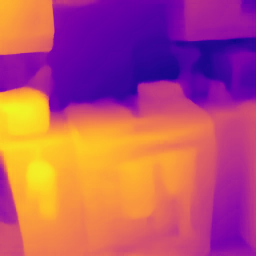} & 
        \includegraphics[width=0.22\textwidth]{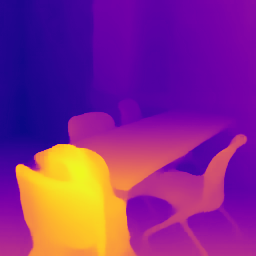} & \includegraphics[width=0.22\textwidth]{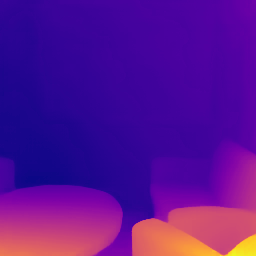} 
        \\
        \normalsize Pred Depth (b0) & \normalsize Pred Depth (b1) & \normalsize Pred Depth (b2) & \normalsize Pred Depth (b3) \\
    \end{tabular}
    \vspace{-3pt}
    \caption{Training logs at last logging step on ScanNet \cite{dai2017scannet} training set. Columns show samples and results of mini-batch ones b0, b1, b2, and b3. For the training logs, we show the color maps of the ground truth depths and predictions in the inverse space (i.e., disparity), so as to better align with the training loss calculated on the inverse depth domain.
    }
    \label{fig:supp_train_logs_A}
\end{figure*}

\begin{figure*}[htp]
    \centering
    \renewcommand{\tabcolsep}{0.2pt}
    \scriptsize
    \begin{tabular}{cccc}
      \includegraphics[width=0.24\textwidth]{train_logs/Step_99609_exp77A/img_0_bs0.png} &  \includegraphics[width=0.24\textwidth]{train_logs/Step_99609_exp77A/img_0_bs1.png} & 
        \includegraphics[width=0.24\textwidth]{train_logs/Step_99609_exp77A/img_0_bs2.png} & \includegraphics[width=0.24\textwidth]{train_logs/Step_99609_exp77A/img_0_bs3.png} 
        \\
        \normalsize Ref $I_0$ (b0) & \normalsize Ref $I_0$ (b1) & \normalsize Ref $I_0$ (b2) & \normalsize Ref $I_0$ (b3) \\
        
      \includegraphics[width=0.24\textwidth]{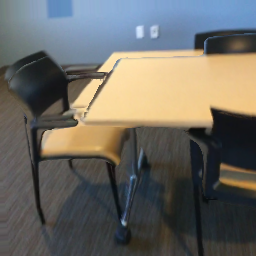} &  \includegraphics[width=0.24\textwidth]{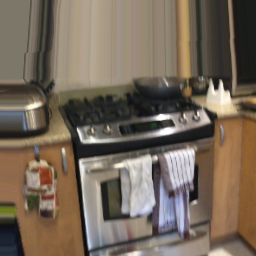} & 
        \includegraphics[width=0.24\textwidth]{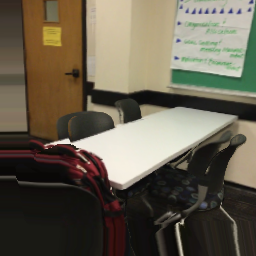} & \includegraphics[width=0.24\textwidth]{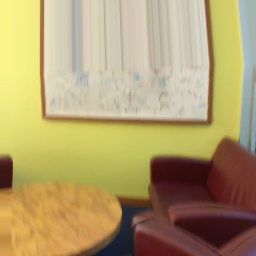} 
        \\
        \normalsize Recon Ref $\tilde{I}_{0\leftarrow{1}}$ (b0) & \normalsize Recon Ref $\tilde{I}_{0\leftarrow{1}}$ (b1) & \normalsize Recon Ref $\tilde{I}_{0\leftarrow{1}}$ (b2) & \normalsize Recon Ref $\tilde{I}_{0\leftarrow{1}}$ (b3) \\
        
     \includegraphics[width=0.24\textwidth]{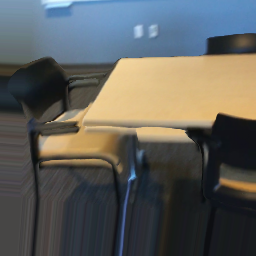} &  \includegraphics[width=0.24\textwidth]{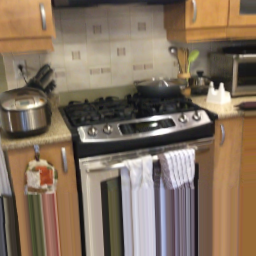} & 
        \includegraphics[width=0.24\textwidth]{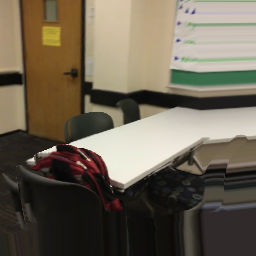} & \includegraphics[width=0.24\textwidth]{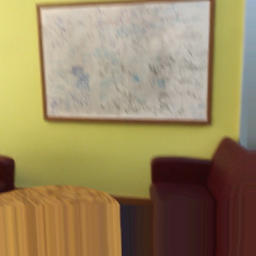} 
        \\
        \normalsize Recon Ref $\tilde{I}_{0\leftarrow{2}}$ (b0) & \normalsize Recon Ref $\tilde{I}_{0\leftarrow{2}}$ (b1) & \normalsize Recon Ref $\tilde{I}_{0\leftarrow{2}}$ (b2) & \normalsize Recon Ref $\tilde{I}_{0\leftarrow{2}}$ (b3) \\

    \end{tabular}
    \caption{Residual pose training. The top row shows the reference images, and the bottom two rows show the reconstructed images of the reference view by warping the source images with the updated poses and predicted depth map of the reference view. 
    }
    \label{fig:supp_train_logs_B}
\end{figure*}

\vspace{5pt}
\noindent \textbf{Network Training and Log Summary.} \,\, Our network is trained from scratch on the ScanNet training set (with 279,494 samples). It takes around 2 days on 4 NVIDIA RTX A6000 GPUs for up to 20 epochs of training. The GRU iteration number is set to 12 for training. The total batch size is 32 (\ie, 8 per GPU). Training image size is 256$\times$256. We show the log summary of network training at the last logging step (i.e., step=99,609). From the top to bottom, Fig.~\ref{fig:supp_train_logs_A} shows a batch of input samples (batch size = 4 for logging), including reference images $I_0$ and two source images $I_1$ and $I_2$, the ground truth depth maps and our depth predictions. The residual pose net is supervised by the photometric loss 
as shown in Fig.~\ref{fig:supp_train_logs_B}. We do one epoch of warmup training only for the residual pose net with other layers frozen.

\begin{figure*}[htp]
    \centering
    \renewcommand{\tabcolsep}{0.2pt}
    \scriptsize
    \begin{tabular}{cccc}
    \includegraphics[width=0.19\textwidth]{train_logs/Step_99609_exp77A/img_0_bs0.png} &  \includegraphics[width=0.19\textwidth]{train_logs/Step_99609_exp77A/img_0_bs1.png} & 
        \includegraphics[width=0.19\textwidth]{train_logs/Step_99609_exp77A/img_0_bs2.png} & \includegraphics[width=0.19\textwidth]{train_logs/Step_99609_exp77A/img_0_bs3.png} 
        \\
        \normalsize Ref $I_0$ (b0) & \normalsize Ref $I_0$ (b1) & \normalsize Ref $I_0$ (b2) & \normalsize Ref $I_0$ (b3) \\
        
        \includegraphics[width=0.19\textwidth]{train_logs/Step_99609_exp77A/gt_disp_bs0.png} &  \includegraphics[width=0.19\textwidth]{train_logs/Step_99609_exp77A/gt_disp_bs1.png} & 
        \includegraphics[width=0.19\textwidth]{train_logs/Step_99609_exp77A/gt_disp_bs2.png} & \includegraphics[width=0.19\textwidth]{train_logs/Step_99609_exp77A/gt_disp_bs3.png} 
        \\
        
        \normalsize GT Depth (b0) & \normalsize GT Depth (b1) & \normalsize GT Depth (b2) & \normalsize GT Depth (b3) \\
\includegraphics[width=0.19\textwidth]{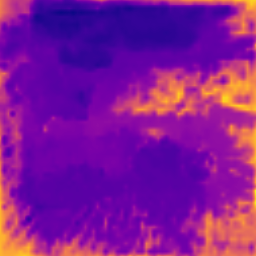} &  \includegraphics[width=0.19\textwidth]{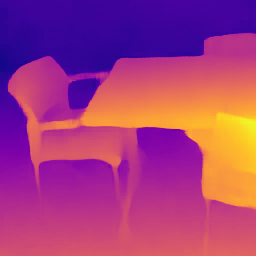} & 
        \includegraphics[width=0.19\textwidth]{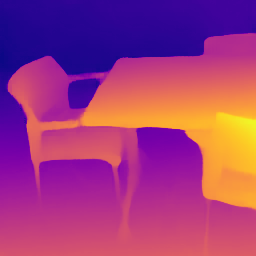} & \includegraphics[width=0.19\textwidth]{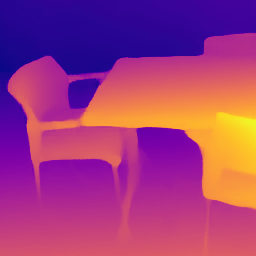} 
        \\
        \normalsize GRU Itr-0 (b0) & \normalsize GRU Itr-4 (b0) & \normalsize GRU Itr-8 (b0) & \normalsize GRU Itr-12 (b0) \\
        
    \includegraphics[width=0.19\textwidth]{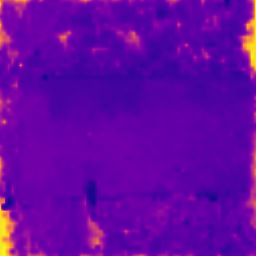} &  \includegraphics[width=0.19\textwidth]{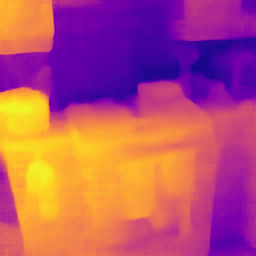} & 
        \includegraphics[width=0.19\textwidth]{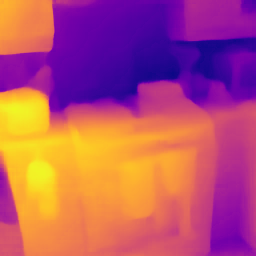} & \includegraphics[width=0.19\textwidth]{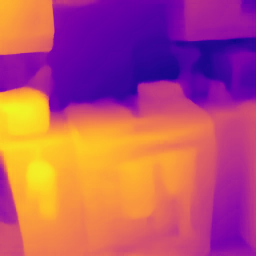} 
        \\
        \normalsize GRU Itr-0 (b1) & \normalsize GRU Itr-4 (b1) & \normalsize GRU Itr-8 (b1) & \normalsize GRU Itr-12 (b1) \\
    
    \includegraphics[width=0.19\textwidth]{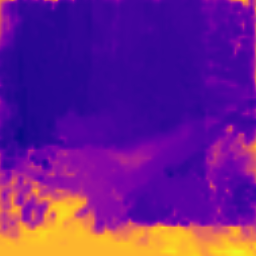} &  \includegraphics[width=0.19\textwidth]{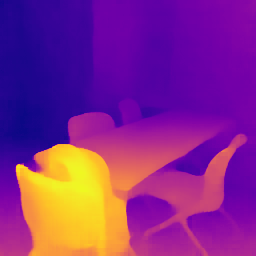} & 
        \includegraphics[width=0.19\textwidth]{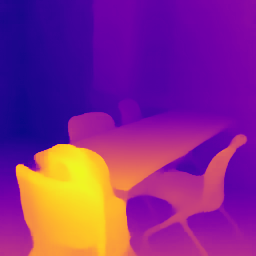} & \includegraphics[width=0.19\textwidth]{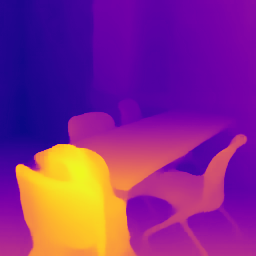} 
        \\
        \normalsize GRU Itr-0 (b2) & \normalsize GRU Itr-4 (b2) & \normalsize GRU Itr-8 (b2) & \normalsize GRU Itr-12 (b2) \\
    
\includegraphics[width=0.19\textwidth]{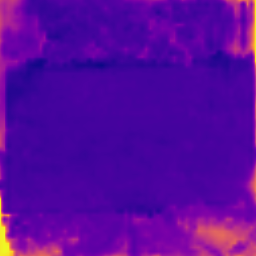} &  \includegraphics[width=0.19\textwidth]{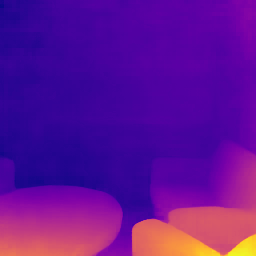} & 
        \includegraphics[width=0.19\textwidth]{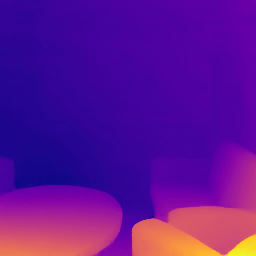} & \includegraphics[width=0.19\textwidth]{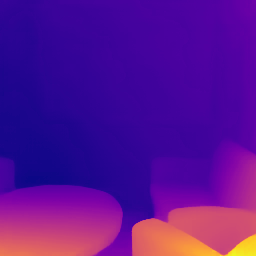} 
        \\
        \normalsize GRU Itr-0 (b3) & \normalsize GRU Itr-4 (b3) & \normalsize GRU Itr-8 (b3) & \normalsize GRU Itr-12 (b3) \\
        
\end{tabular}
    \vspace{-3pt}
    \caption{Iterative depth estimation from GRU layers. The bottom 4 rows show the depth predictions at iteration step $t=0,4,8,12$ for each batch sample (b1, b2, b3 and b4).
    }
    \label{fig:supp_train_logs_C}
\end{figure*}

\vspace{5pt}
\noindent \textbf{GRU Iterative Updates.} \,\, Fig.~\ref{fig:supp_train_logs_C} 
illustrates the iterative estimation of depth maps. For better visualization, we put the reference images and the ground truth depths on the first two rows. The bottom 4 rows show the depth predictions at iteration step $t=0,4,8,12$ for each batch sample. \textit{Itr-0} means the softargmin-start we introduced to accelerate the GRU training and convergence. We can see the depth maps are progressively improved within $T$ iterations (here $T=12$ in network training for the trade-off between the memory consumption and depth accuracy).

\vspace{5pt}
\noindent \textbf{Network Inference.} \,\, For inference, we set the GRU iteration number as $T=24$ by default, and we also ablate other values of $T$ in the main paper. The input image is in 320$\times$256 resolution, and it is upsampled to 640$\times$320 for ScanNet benchmark evaluation and cross-dataset generalization. The GPU memory consumption is 2088MiB from \textit{nvidia-smi}, and runtime in inference mode is 8.6 fps when processing frames with dimension 320 $\times$256.

\section{Additional Ablation Studies}
\begin{table}[!htb]
\centering
\scalebox{0.9}{
        \begin{tabular}{l | c c c}
        \Xhline{2\arrayrulewidth}
        Sampling & abs-rel & abs & $\delta <$ 1.25  \\
	\hline
	s10 (base) & 0.0885 & 0.1605 & 0.9211 \\
	key (base) & 0.0838 & 0.1598 & 0.9277 \\
    
    \hline
    s10 (+pose) & 0.0827 & 0.1523 & 0.9277 \\
    key (+pose) & 0.0789 & 0.1531 & 0.9339 \\
    
    \hline
    s10 (+pose,atten) & \underline{0.0747} & \underline{0.1392} & \underline{0.9382} \\
    key (+pose,atten) &  \textbf{0.0697} & \textbf{0.1348} & \textbf{0.9472} \\
	\Xhline{2\arrayrulewidth}
	
    \end{tabular}
    }
\vspace{2pt}

    \caption{Frame sampling comparison. The results are evaluated on the ScanNet test set~\cite{dai2017scannet}.} 
\label{tab:supp-frame-sampling}
\end{table}
We introduce more ablation studies to verify our design.

\vspace{5pt}
\noindent \textbf{Frame Sampling:} \,\, We compare the simple view selection strategy (i.e., sampling by every 10 frames), with the heuristics introduced in \cite{deepvideomvs2021}. Tab.~\ref{tab:supp-frame-sampling} shows that our methods can be further improved when the selected views have more overlapping and the baselines between them are suitable. Our(+pose,atten) even with simple strategy outperforms other variants with heuristic sampling, and so are our(+pose) vs our(base), verifying the effectiveness of each module. 

\vspace{5pt}
\noindent \textbf{Different Depth Binning.} \,\, When implementing plane-sweep stereo \cite{space-sweep-collins, gallup2007real} to construct the cost volume, we need to sample $M_0$=64 plane hypotheses. In our main experiments, we use the inverse depth bins, \ie, the plane hypotheses are uniformly sampled in the inverse depth space, s.t. $ 1/d \sim U(d_{\text{min}}, d_{\text{max}})$. Here we set $d_{min}$=0.25 and $d_{max}$=20 meters for indoor scenes (e.g., ScanNet \cite{dai2017scannet}). We also test linear depth bins, i.e., $d \sim U(d_{\text{min}}, d_{\text{max}})$, and hand-crafted depth bins by calculating the depth distribution on ScanNet. But we found that inverse depth binning achieves the best results, as we reported in the main paper. We also test adaptive depth bins as in \cite{bhat2021adabins}, where the depth bins are dynamically generated upon the global feature learned by a transformer layer. For our(+pose) variant, adaptive depth bins lead to marginal improvement than the inverse depth bins. However, for our(+pose,atten) variant, adaptive depth bins hinder the depth accuracy.

\section{Qualitative Results}

\begin{figure}[htp]
    \centering
    \renewcommand{\tabcolsep}{0.2pt}
    \scriptsize
      \includegraphics[width=0.48\textwidth]{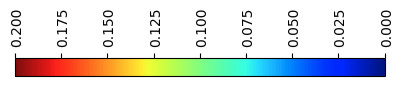}
    \caption{Color scale used for all \textit{abs} error in depth maps in the supplementary material. }
    \label{fig:err_clr_map}
\end{figure}

\begin{figure*}[htp]
    \centering
    \renewcommand{\tabcolsep}{0.2pt}
    \scriptsize
    \begin{tabular}{cccc}
      \includegraphics[width=0.195\textwidth]{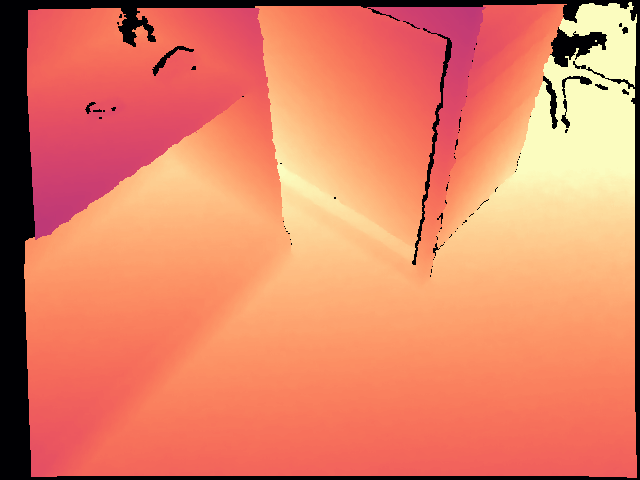} &  \includegraphics[width=0.195\textwidth]{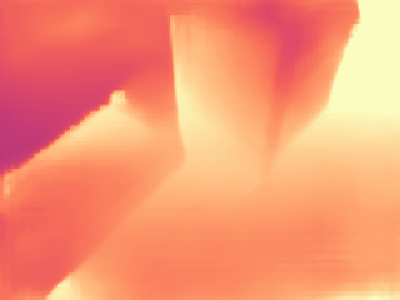} & 
        \includegraphics[width=0.195\textwidth]{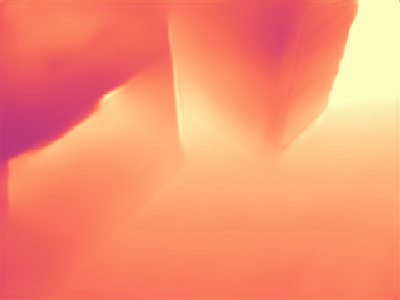} & \includegraphics[width=0.195\textwidth]{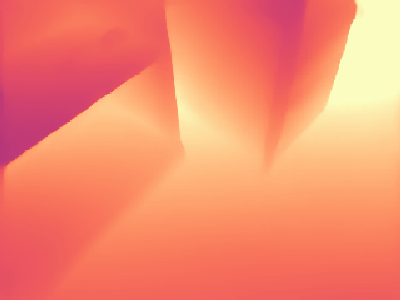} 
        \\
        \includegraphics[width=0.195\textwidth]{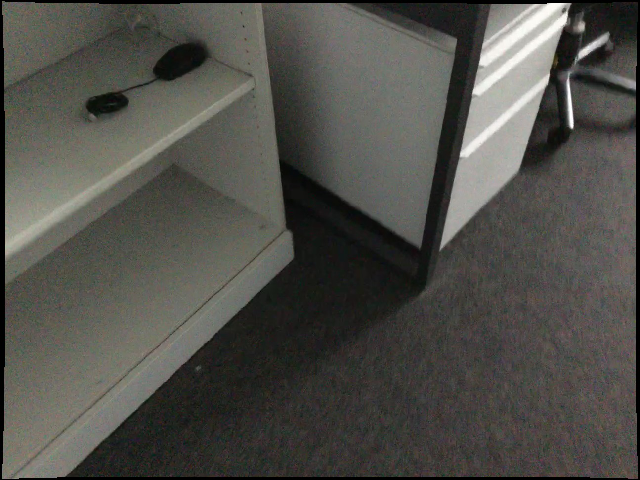}
         &  
         \begin{overpic}[width=0.195\textwidth]{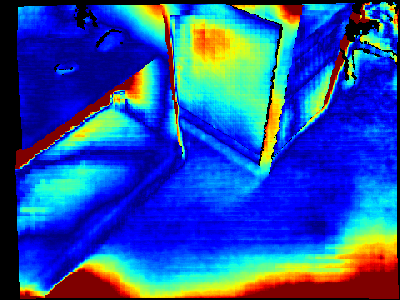}
        \put (10,60) {$\displaystyle\textcolor{white}{\textbf{0.0660}}$}
        \end{overpic} & 
        \begin{overpic}[width=0.195\textwidth]{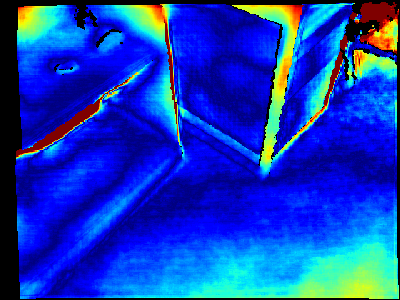}
        \put (10,60) {$\displaystyle\textcolor{white}{\textbf{0.0421}}$}
        \end{overpic} &
        \begin{overpic}[width=0.195\textwidth]{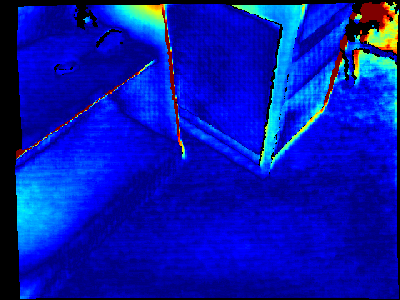}
        \put (10,60) {$\displaystyle\textcolor{white}{\textbf{0.0245}}$}
        \end{overpic}
        \\
        \includegraphics[width=0.195\textwidth]{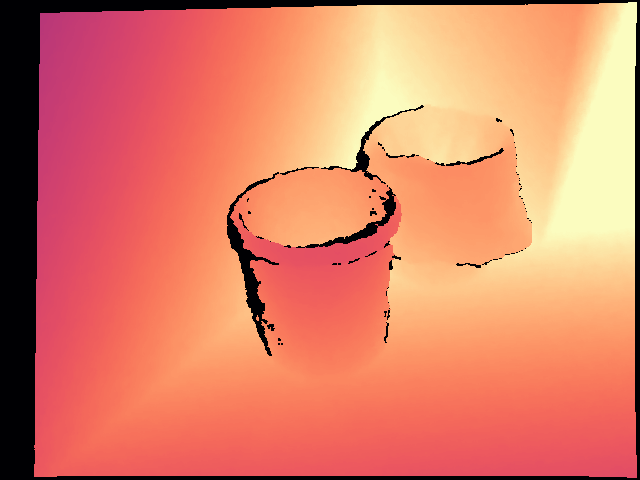} &  \includegraphics[width=0.195\textwidth]{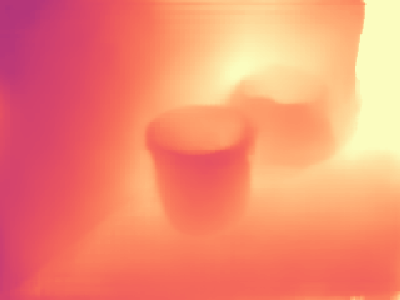} & 
        \includegraphics[width=0.195\textwidth]{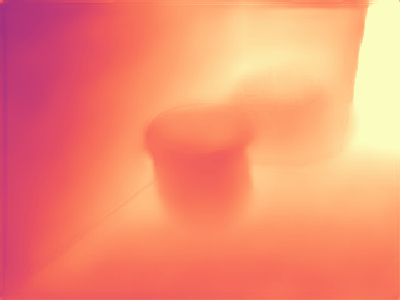} & \includegraphics[width=0.195\textwidth]{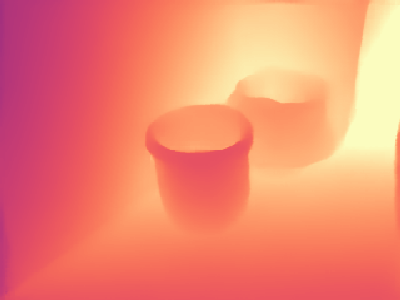} 
        \\
        \includegraphics[width=0.195\textwidth]{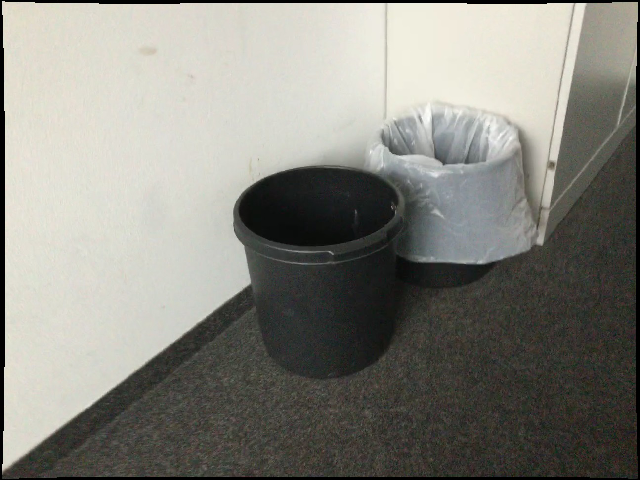}
         &  
         \begin{overpic}[width=0.195\textwidth]{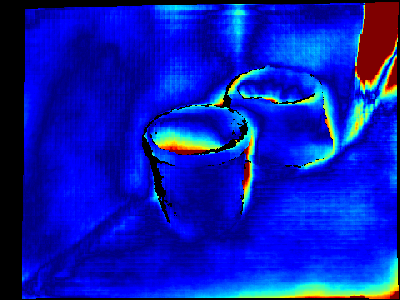}
        \put (10,60) {$\displaystyle\textcolor{white}{\textbf{0.0362}}$}
        \end{overpic} & 
        \begin{overpic}[width=0.195\textwidth]{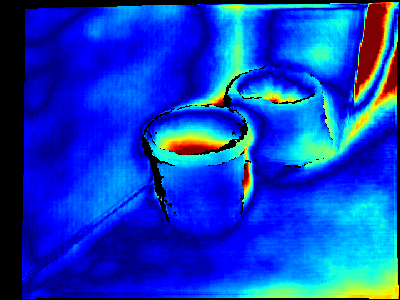}
        \put (10,60) {$\displaystyle\textcolor{white}{\textbf{0.0417}}$}
        \end{overpic} &
        \begin{overpic}[width=0.195\textwidth]{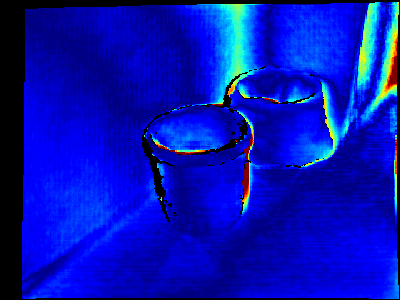}
        \put (10,60) {$\displaystyle\textcolor{white}{\textbf{0.0210}}$}
        \end{overpic}
        \\

        \includegraphics[width=0.195\textwidth]{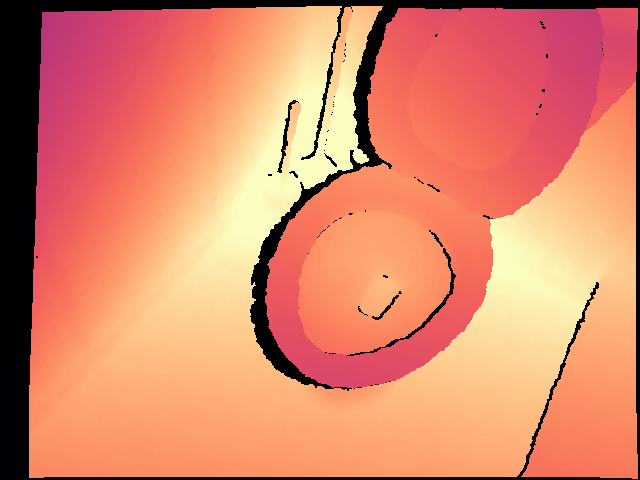} &  \includegraphics[width=0.195\textwidth]{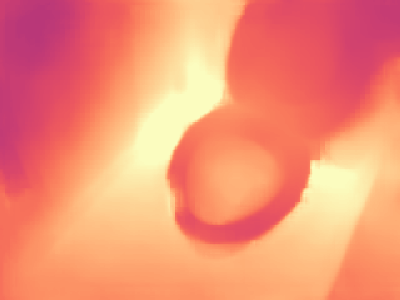} & 
        \includegraphics[width=0.195\textwidth]{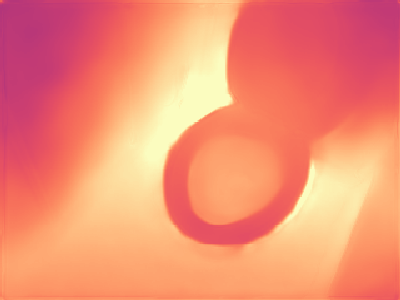} & \includegraphics[width=0.195\textwidth]{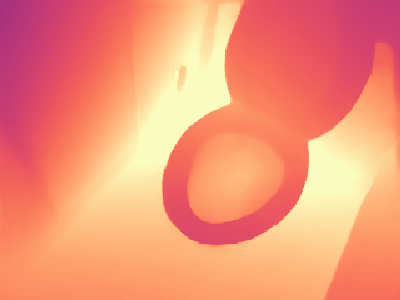} 
        \\
        \includegraphics[width=0.195\textwidth]{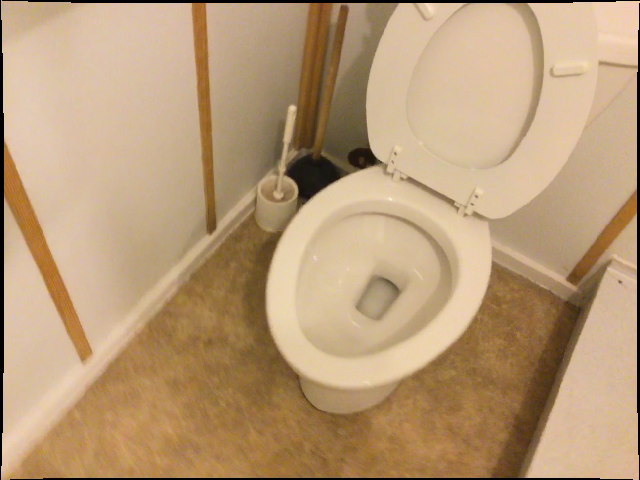}
         &  
         \begin{overpic}[width=0.195\textwidth]{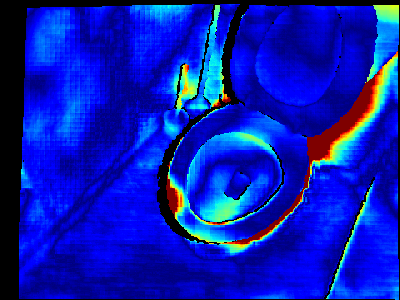}
        \put (10,60) {$\displaystyle\textcolor{white}{\textbf{0.0297}}$}
        \end{overpic} & 
        \begin{overpic}[width=0.195\textwidth]{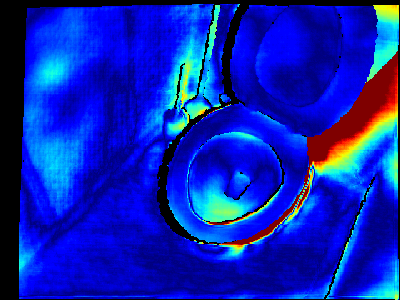}
        \put (10,60) {$\displaystyle\textcolor{white}{\textbf{0.0340}}$}
        \end{overpic} &
        \begin{overpic}[width=0.195\textwidth]{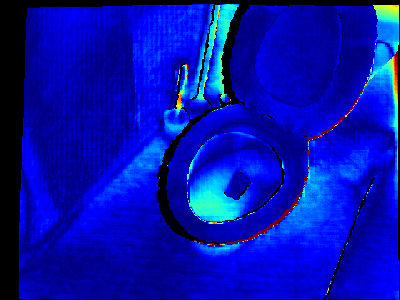}
        \put (10,60) {$\displaystyle\textcolor{white}{\textbf{0.0189}}$}
        \end{overpic}
        \\

        \includegraphics[width=0.195\textwidth]{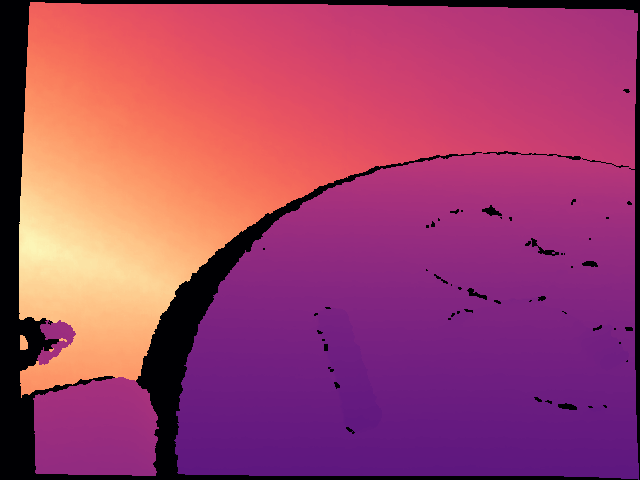} &  \includegraphics[width=0.195\textwidth]{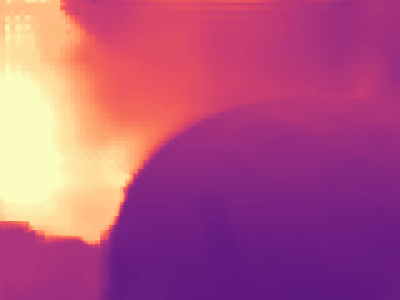} & 
        \includegraphics[width=0.195\textwidth]{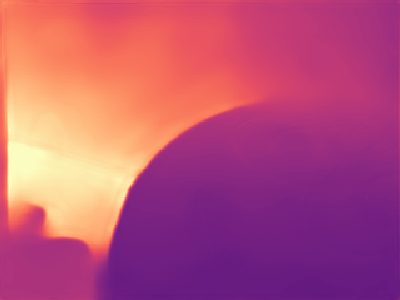} & \includegraphics[width=0.195\textwidth]{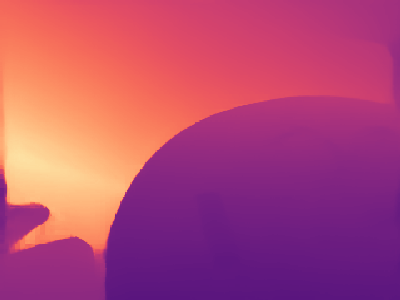} 
        \\
        \includegraphics[width=0.195\textwidth]{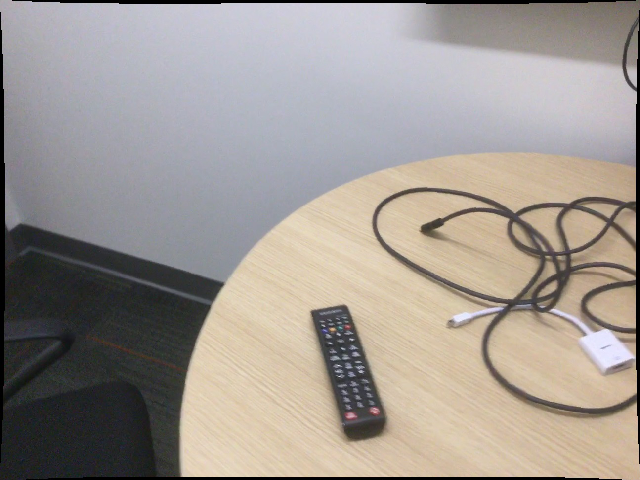}
         &  
         \begin{overpic}[width=0.195\textwidth]{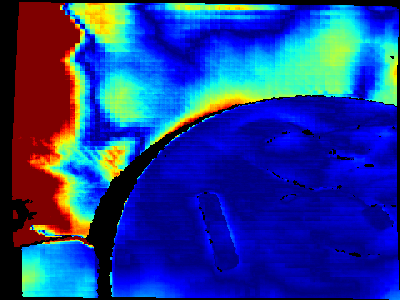}
        \put (25,60) {$\displaystyle\textcolor{white}{\textbf{0.0662}}$}
        \end{overpic} & 
        \begin{overpic}[width=0.195\textwidth]{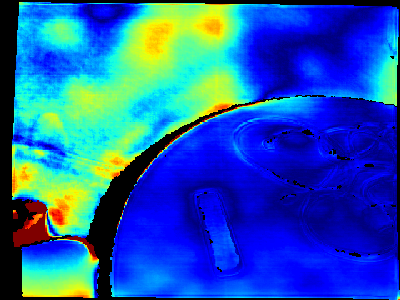}
        \put (10,60) {$\displaystyle\textcolor{white}{\textbf{0.0520}}$}
        \end{overpic} &
        \begin{overpic}[width=0.195\textwidth]{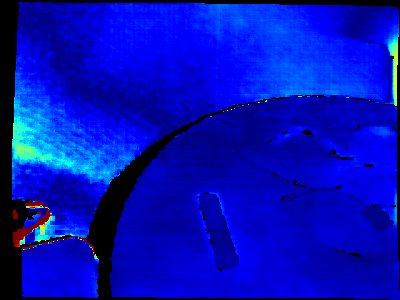}
        \put (10,60) {$\displaystyle\textcolor{white}{\textbf{0.0217}}$}
        \end{overpic}
        \\
        \normalsize GT & \normalsize IterMVS & \normalsize PairNet & \normalsize Ours \\
    
    \end{tabular}
    \vspace{-3pt}
    \caption{Qualitative results on ScanNet \cite{dai2017scannet} test set. Every two rows show depth maps (top) and error maps (bottom) for a sample. The leftmost column shows ground truth depths and reference images. Others columns are the depth predictions and error maps, by IterMVS \cite{itermvs_wang2021}, PairNet \cite{deepvideomvs2021} and ours, respectively. The abs-err errors (in meters) are imposed on the depth maps for comparison.
    }
    \label{fig:supp_qualitatives_scannet}
\end{figure*}

\noindent \textbf{Depth and Error Maps.} \, More qualitative results of depth maps and error maps on the ScanNet test set~\cite{dai2017scannet} are shown in Fig.~\ref{fig:supp_qualitatives_scannet}. The samples shown here are {\myqcrfont scene0711\_00/001050.png}, {\myqcrfont scene0711\_00/002530.png}, {\myqcrfont scene0727\_00/001260.png}, and {\myqcrfont scene0769\_00/000720.png}. The error maps contain the absolute errors \textit{abs} in depth. For the ground truth depth maps and the error maps, invalid regions (i.e., without ground truth depth annotation) are filled in black. The color maps of the ground truth depths and predictions are shown in the depth space (i.e., not in disparity space). The \textit{abs} errors (in meters) are superimposed on the error maps for better comparison. The corresponding color bar to visualize the error maps is shown in Fig.~\ref{fig:err_clr_map}. Comparing the depth predictions and the error maps for our method and the baseline IterMVS \cite{itermvs_wang2021} and baseline PairNet \cite{deepvideomvs2021}, our method predicts more accurate estimates, especially in the challenging regions, \eg, the boundary, the ground, the white wall, and the round desk. 

\begin{figure*}[htp]
    \centering
    \renewcommand{\tabcolsep}{0.2pt}
    \scriptsize
    \begin{tabular}{ccccc}
      \includegraphics[width=0.20\textwidth]{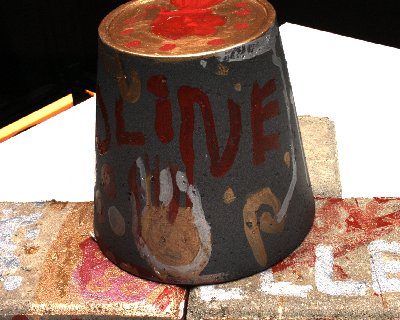} & \includegraphics[width=0.20\textwidth]{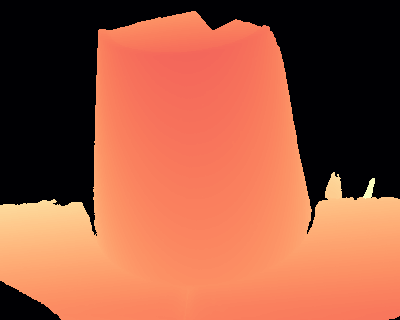} & 
      
      \begin{overpic}[width=0.20\textwidth]{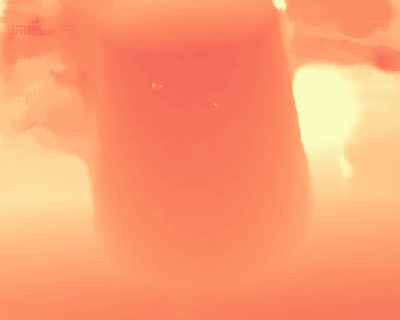}
        \put (7,65) {$\displaystyle\textcolor{white}{\textbf{4.45}}$}
        \end{overpic} &
        		
        \begin{overpic}[width=0.20\textwidth]{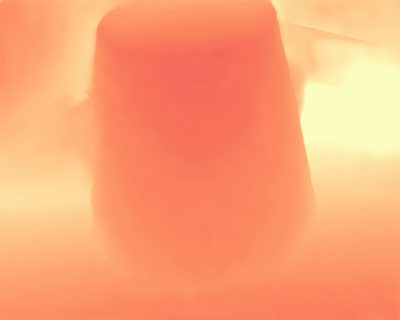}
        \put (7, 65) {$\displaystyle\textcolor{white}{\textbf{6.59}}$}
        \end{overpic} &
        
        \begin{overpic}[width=0.20\textwidth]{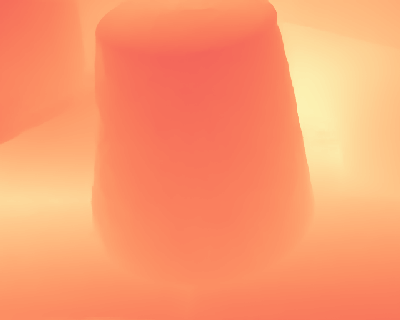}
        \put (7, 65) {$\displaystyle\textcolor{white}{\textbf{2.18}}$}
        \end{overpic}
        
        \\
        
      \includegraphics[width=0.20\textwidth]{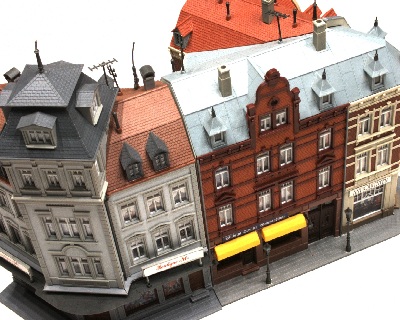} & \includegraphics[width=0.20\textwidth]{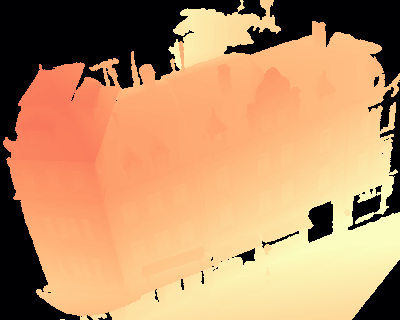} & 
      		
      \begin{overpic}[width=0.20\textwidth]{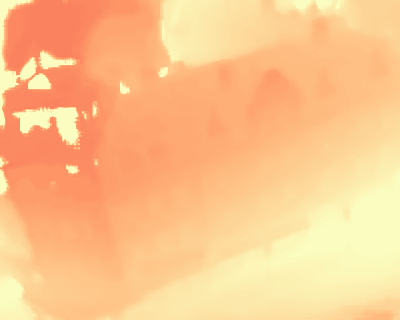}
        \put (7, 65) {$\displaystyle\textcolor{white}{\textbf{14.24}}$}
        \end{overpic} &
        
        \begin{overpic}[width=0.20\textwidth]{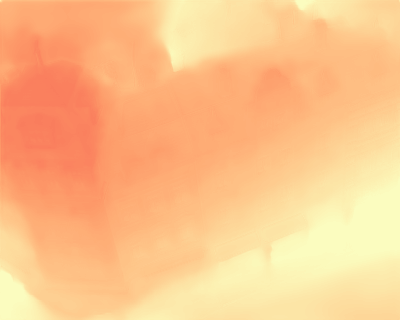}
        \put (7, 65) {$\displaystyle\textcolor{white}{\textbf{6.28}}$}
        \end{overpic} &
        
        \begin{overpic}[width=0.20\textwidth]{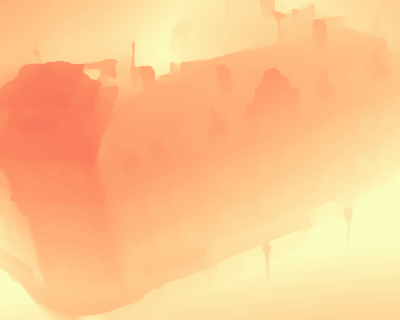}
        \put (7, 65) {$\displaystyle\textcolor{white}{\textbf{3.28}}$}
        \end{overpic}
        
        \\
        
      \includegraphics[width=0.20\textwidth]{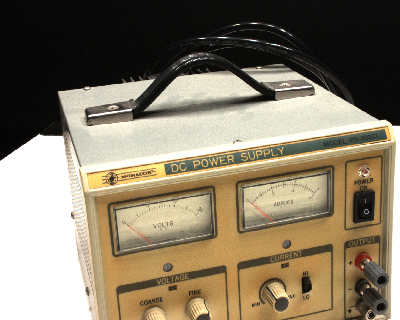} & \includegraphics[width=0.20\textwidth]{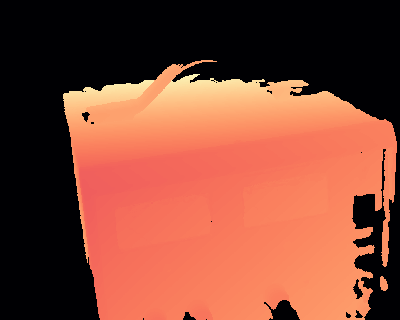} & 
      
      \begin{overpic}[width=0.20\textwidth]{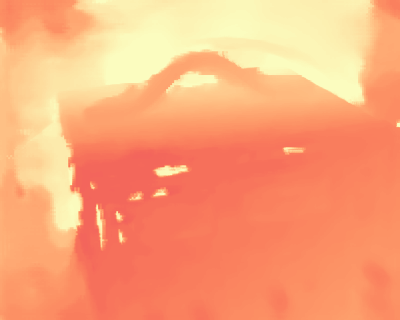}
        \put (7, 65) {$\displaystyle\textcolor{white}{\textbf{12.12}}$}
        \end{overpic} &
        
        \begin{overpic}[width=0.20\textwidth]{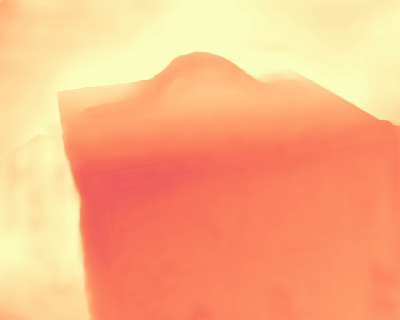}
        \put (7, 65) {$\displaystyle\textcolor{white}{\textbf{12.91}}$}
        \end{overpic} &
        
        \begin{overpic}[width=0.20\textwidth]{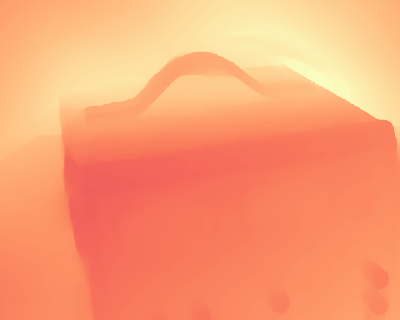}
        \put (7, 65) {$\displaystyle\textcolor{white}{\textbf{5.27}}$}
        \end{overpic}
        		
        \\

        \\

      \includegraphics[width=0.20\textwidth]{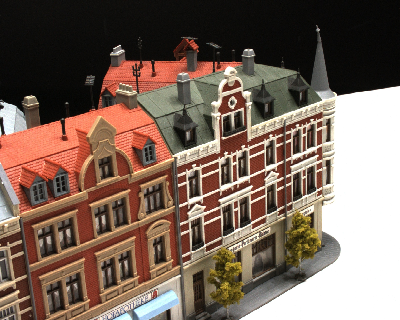} & \includegraphics[width=0.20\textwidth]{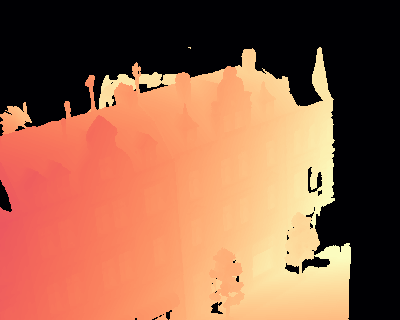} & 
      
      \begin{overpic}[width=0.20\textwidth]{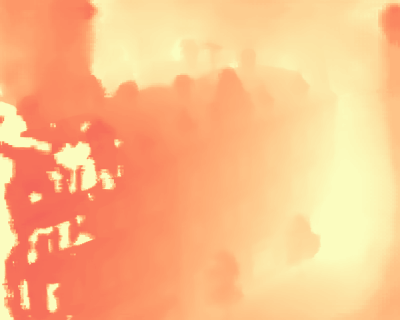}
        \put (7, 65) {$\displaystyle\textcolor{white}{\textbf{40.10}}$}
        \end{overpic} &
        
        \begin{overpic}[width=0.20\textwidth]{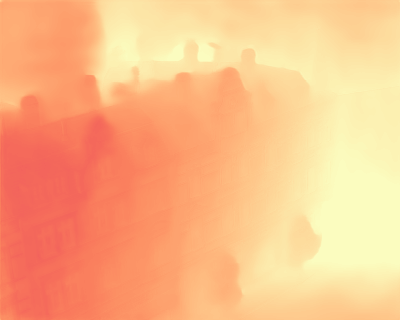}
        \put (7, 65) {$\displaystyle\textcolor{white}{\textbf{12.19}}$}
        \end{overpic} &
        
        \begin{overpic}[width=0.20\textwidth]{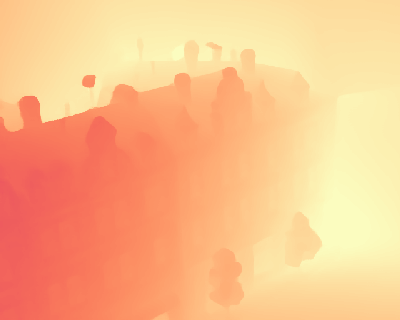}
        \put (7, 65) {$\displaystyle\textcolor{white}{\textbf{5.31}}$}
        \end{overpic}
        	
        \\
        
      \includegraphics[width=0.20\textwidth]{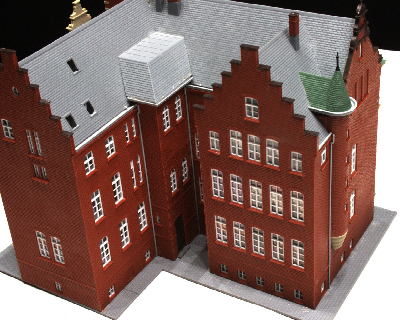} & \includegraphics[width=0.20\textwidth]{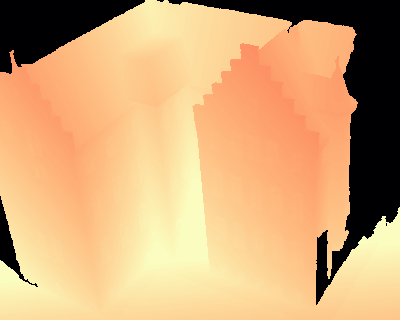} & 
      
      \begin{overpic}[width=0.20\textwidth]{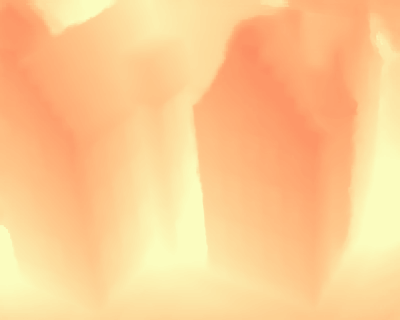}
        \put (7, 65) {$\displaystyle\textcolor{white}{\textbf{5.30}}$}
        \end{overpic} &
        
        \begin{overpic}[width=0.20\textwidth]{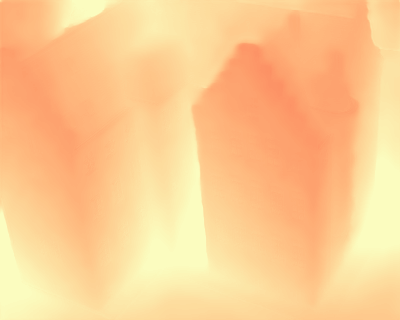}
        \put (7, 65) {$\displaystyle\textcolor{white}{\textbf{6.09}}$}
        \end{overpic} &
        
        \begin{overpic}[width=0.20\textwidth]{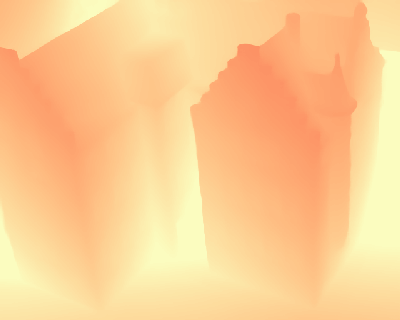}
        \put (7, 65) {$\displaystyle\textcolor{white}{\textbf{3.35}}$}
        \end{overpic}
		
        \\
        
      \includegraphics[width=0.20\textwidth]{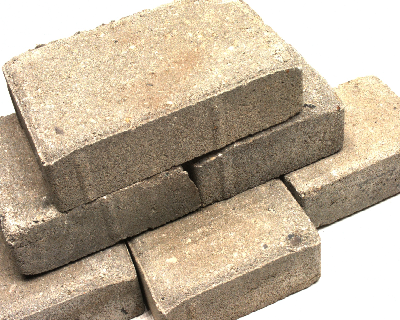} & \includegraphics[width=0.20\textwidth]{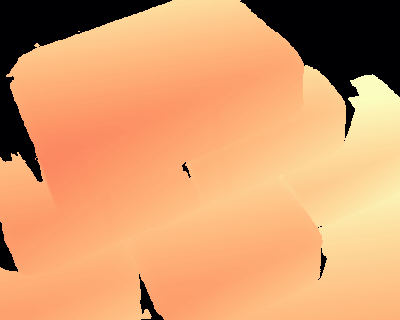} & 
      
      \begin{overpic}[width=0.20\textwidth]{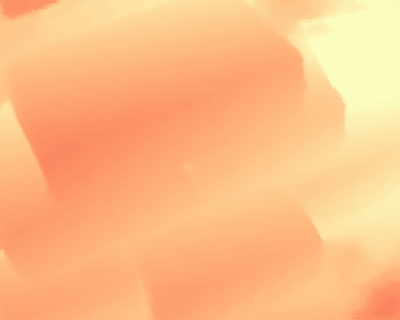}
        \put (7, 65) {$\displaystyle\textcolor{white}{\textbf{4.64}}$}
        \end{overpic} &
        
        \begin{overpic}[width=0.20\textwidth]{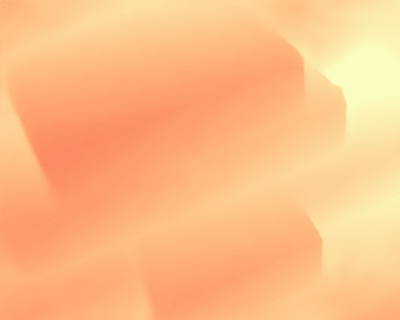}
        \put (7, 65) {$\displaystyle\textcolor{white}{\textbf{4.86}}$}
        \end{overpic} &
        
        \begin{overpic}[width=0.20\textwidth]{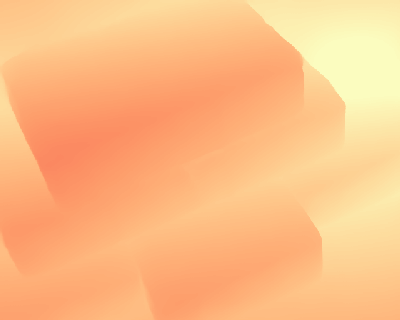}
        \put (7, 65) {$\displaystyle\textcolor{white}{\textbf{2.04}}$}
        \end{overpic}
        		
        \\
        
      \includegraphics[width=0.20\textwidth]{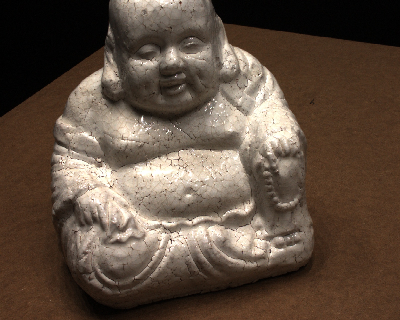} & \includegraphics[width=0.20\textwidth]{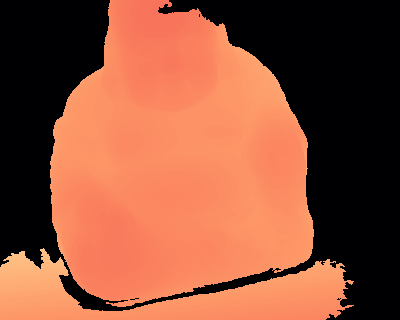} & 
      
      \begin{overpic}[width=0.20\textwidth]{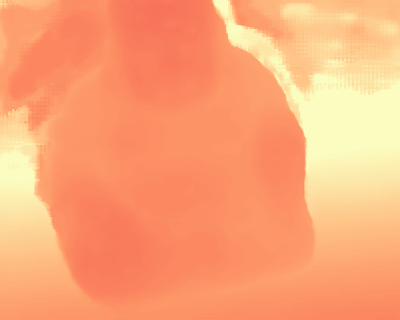}
        \put (7, 65) {$\displaystyle\textcolor{white}{\textbf{4.10}}$}
        \end{overpic} &
        
        \begin{overpic}[width=0.20\textwidth]{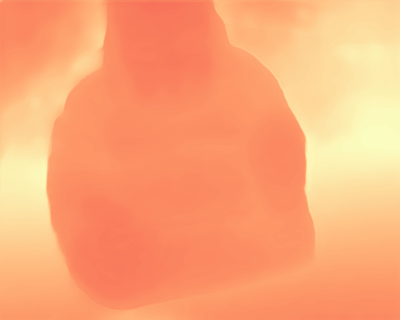}
        \put (7, 65) {$\displaystyle\textcolor{white}{\textbf{5.52}}$}
        \end{overpic} &
        
        \begin{overpic}[width=0.20\textwidth]{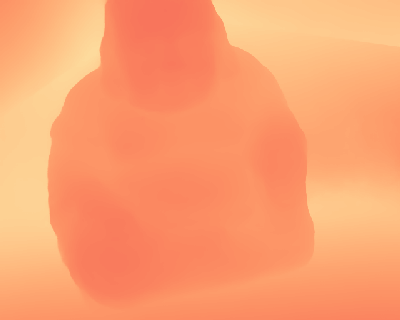}
        \put (7, 65) {$\displaystyle\textcolor{white}{\textbf{1.97}}$}
        \end{overpic}
        	
        \\
        
        \\
        \normalsize Ref & \normalsize GT & \normalsize IterMVS & \normalsize PairNet & \normalsize Ours \\
    
    \end{tabular}
    \vspace{-3pt}
   \caption{Cross-dataset generalization qualitative results on DTU \cite{dtu_jensen2014large} trained on ScanNet. Columns from left to right show reference image, ground truth depth, and the estimated depth for baseline IterMVS \cite{itermvs_wang2021}, PairNet \cite{deepvideomvs2021} and our method, respectively. Our method outperforms IterMVS visibly, and on par with PairNet. The abs-err errors (in millimeters) are imposed on the depth maps for comparison. 
    }
    \label{fig:supp_qualitatives_scan2dtu_v2}
\end{figure*}

\vspace{5pt}
\noindent \textbf{Cross-Dataset Generalization from ScanNet to DTU} \,\, Fig.~\ref{fig:supp_qualitatives_scan2dtu_v2} shows the depth maps of DTU dataset when generalized from ScanNet without fine-tuning, and our method outperforms IterMVS visibly, and on par with PairNet. 

\begin{figure*}[htp]
    \centering
    \renewcommand{\tabcolsep}{0.2pt}
    \scriptsize
      \includegraphics[width=0.9\textwidth]{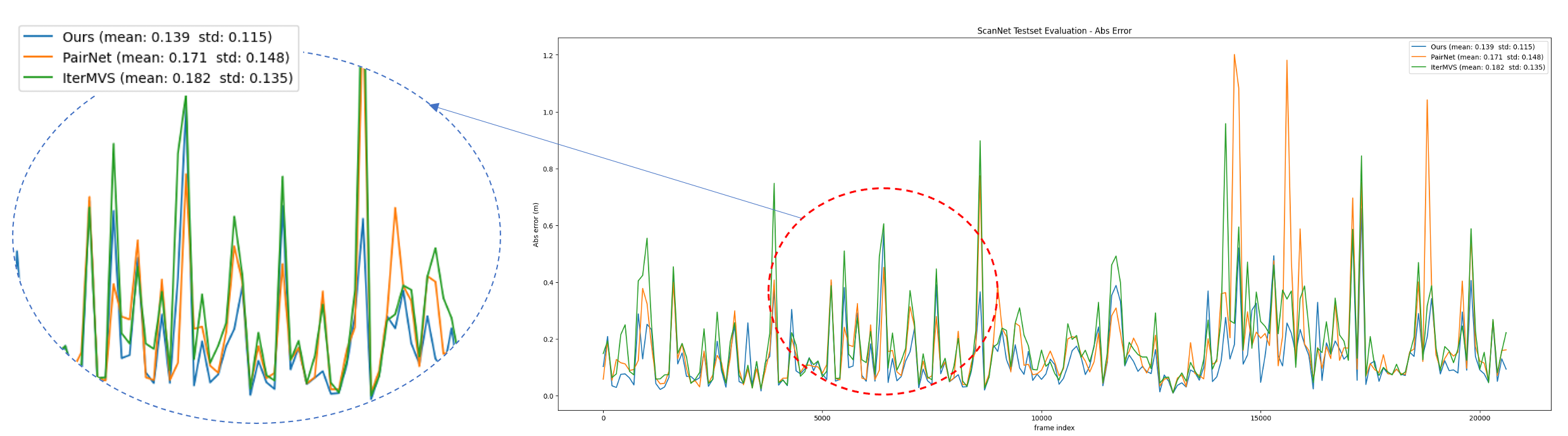}
    
    \caption{Absolute error metric curves evaluated on all the frames of ScanNet  \cite{dai2017scannet} test set, for our method and baselines IterMVS \cite{itermvs_wang2021} and PairNet \cite{deepvideomvs2021}. Please enlarge the figures to better view the metrics and legends displayed in the top-left corner.
    }
    \label{fig:abs_curve_scannet}
\end{figure*}

The ScanNet test set in our experiments contains 20,668 samples. As shown in Fig. \ref{fig:abs_curve_scannet}, we report \textit{abs} error curves (by plotting values in meters every 100 frames, out of those 20,668 samples) to reflect the distribution of the errors. We also compare the mean and standard deviation to reflect the overall performance of our method versus the baselines: mean error 0.139 (our) $<$ 0.171 (PairNet) $<$ 0.182 (IterMVS), and standard deviation 0.115 (our) $<$ 0.135 (IterMVS) $<$ 0.148 (PairNet), showing that our method consistently outperforms the baselines with smaller average and lower standard deviation.

\section{Quantitative Metrics}

We use the metrics defined in \cite{eigen2014depth}, including mean absolute error ({\myqcrfont abs}), mean absolute relative error ({\myqcrfont abs-rel}), squared relative error ({\myqcrfont sq-rel}), RMSE in linear ({\myqcrfont rmse}) and log ({\myqcrfont rmse-log}) scales, and inlier ratios under thresholds of 1.25/1.25$^2$/1.25$^3$. For a predicted depth map $y$ and ground truth $y^*$, each with $n$ pixels indexed by $i$, those metrics are formulated as:

\begingroup
\setlength{\abovedisplayskip}{0pt}
\begin{align*}
& \text{{\myqcrfont abs}}:  \frac{1}{n} \sum_{i} | y_i - y_i^*| \\
& \text{{\myqcrfont abs-rel}}:  \frac{1}{n} \sum_{i} | y_i - y_i^*|/ y_i^* \quad
\text{{\myqcrfont sq-rel}}:  \frac{1}{n} \sum_{i} \| y_i - y_i^* \|^2 / y_i^* \\ 
& \text{{\myqcrfont rmse}}: \sqrt{\frac{1}{n} \sum_{i} \| y_i - y_i^* \|^2}  \\
& \text{{\myqcrfont rmse-log}}: \sqrt{\frac{1}{n} \sum_{i} \| \log {y_i} - \log{y_i^*} \|^2}  \\
& \text{inlier {\myqcrfont ratio}} \%  \text{ of } y_i \quad \text{s.t. }  \text{max} \left( \frac{y_i}{y_i^*},   \frac{y_i^*}{y_i} \right) = \delta < 1.25^{i}, \\
& \qquad \qquad \qquad \qquad \text{ where } i=1, 2 \text{ and } 3.
\end{align*}
\endgroup

\end{document}